\documentclass[times]{speauth}

\usepackage{moreverb}
\usepackage{amsmath}
\usepackage{algorithm}
\usepackage{algpseudocode}
\usepackage{booktabs}
\usepackage{tabularx}
\usepackage{multirow}
\usepackage{wrapfig}
\usepackage{graphicx}
\usepackage{subcaption}
\usepackage{url}
\usepackage{verbatim}
\usepackage{tabularx}

\usepackage[superscript,biblabel]{cite}

\newcommand\BibTeX{{\rmfamily B\kern-.05em \textsc{i\kern-.025em b}\kern-.08em
T\kern-.1667em\lower.7ex\hbox{E}\kern-.125emX}}

\begin{document}

\runningheads{Massobrio, Toutouh, Nesmachnow, and Alba}{Infrastructure Deployment in VANETs Using a Parallel MOEA}

\title{Infrastructure Deployment in Vehicular Communication Networks \\Using a Parallel Multiobjective Evolutionary Algorithm}

\author{Renzo Massobrio\affil{1}\corrauth, Jamal Toutouh\affil{2}, Sergio Nesmachnow\affil{1}, Enrique Alba\affil{2}}

\address{%
\affilnum{1}Universidad de la Rep\'ublica, Herrera y Reissig 565, Montevideo, 11300, Uruguay \break
\affilnum{2}Dept. de Lenguajes y Ciencias de la Computación, University of M\'alaga, M\'alaga, 29071, Spain}
%-
\corraddr{Author to whom all correspondence should be addressed: e-mail: renzom@fing.edu.uy}

\begin{abstract}
This article describes the application of a multiobjective evolutionary algorithm for locating roadside infrastructure for vehicular communication networks over realistic urban areas. A multiobjective formulation of the problem is introduced, considering quality-of-service and cost objectives. The experimental analysis is performed over a real map of M\'alaga, using real traffic information and antennas, and scenarios that model different combinations of traffic patterns and applications (text/audio/video) in the communications. The proposed multiobjective evolutionary algorithm computes accurate trade-off solutions, significantly improving over state-of-the-art algorithms previously applied to the problem.
\end{abstract}


\maketitle

\vspace{-6pt}
\section{Introduction}
\label{sec:intro}
\vspace{-2pt}

Vehicular traffic is a major concern in modern cities\cite{Falcocchio2015}. Several problems related to mobility, traffic safety, environment, etc.~can be efficiently solved by applying smart computational methods. In this context, the concept of \textit{smart cities} has emerged as a key issue in modern urbanization. A smart city applies information technologies to enhance quality, performance, and interactivity of urban services and/or to reduce costs and resource consumption. Road traffic management is a specific area that makes use of smart city applications, with the goal of 
improving the management of urban flows and allowing for real time responses to challenges that have great impact on the citizens\cite{Deakin2013}.

A number of smart city solutions are based on \textit{intelligent transport systems} (ITS). 
The main idea behind these systems consist in sharing information about the traffic conditions with road users and authorities. A better informed citizen can take better driving decisions, positively influencing the global traffic safety, efficiency. 

\textit{Vehicular ad hoc networks} (VANETs) emerge as a promising technology to allow continuous data exchange between vehicles equipped with an on-board unit (OBU). Vehicles can also communicate with \textit{roadside unit} (RSU) elements via \textit{direct short range communications} (DSRC).   
Depending on the type of nodes involved, several types of communications can occur in a VANET: vehicle-to-vehicle (V2V) communications, when the vehicles communicate directly with each other, and vehicle-to-infrastructure (V2I) communications, when the vehicles exchange data with RSUs. 

VANETs allow developing a large set of powerful applications to improve road transport experience for both drivers and passengers. 
Typically, these applications are categorized into \textit{safety} and \textit{non-safety} applications. 
The first ones aim at improving road safety and avoiding hazardous situations (road accidents), e.g., cooperative driving and intersection collision avoidance applications.   
Non-safety applications include a collection of different solutions oriented to enhancing the traffic efficiency (e.g., travel times, fuel consumption, CO$_2$ emissions, etc.). These applications also allow improving the  comfort and entertainment of passengers. 

Safety and traffic efficiency applications, such as \textit{Cooperative Vehicle Safety}, gather real-time data from diverse sources (vehicle sensors, information received from other nodes, or both), process it, and disseminate it to the other nodes. Most of these applications rely on periodic message broadcasting or 
beaconing. 
This kind of applications require very short data delivery times (in terms of milliseconds)\cite{Campolo2015}, since
larger delivery times increase the uncertainty on the system and may cause hazardous situations. 
Infotainment VANET applications (e.g., audio or video broadcasting, on-line gaming, etc.) mainly rely on continuous data streams. The real time requirements are lower than for safety and efficiency applications, but they require larger transmission data rates (in terms of tenths of kilobytes per second) to keep the quality of the service provided. 

In this study, we focus on a specific element of VANET architecture, the RSUs, which are devices that are usually installed along the roads on the roadside infrastructure elements, e.g., traffic lights. In addition, they may be fixed along roadside as specific dedicated VANET elements. RSUs include a network interface to exchange information with other VANET nodes through DSRC. 
They may also be equipped with other network interfaces to connect to other networks or to the Internet.
RSUs perform three main functions: i) acting as an information transmitter or receiver in VANET applications, e.g., warning about of the existence of roadworks, accidents, etc.; ii) extending the effective communication range by forwarding data to other VANET nodes (OBUs or RSUs) through multi-hop communications; and iii) providing Internet connectivity to other nodes in the VANET. 

\begin{figure}
\centering
\includegraphics[width=\linewidth,trim={0 40 0 80},clip]{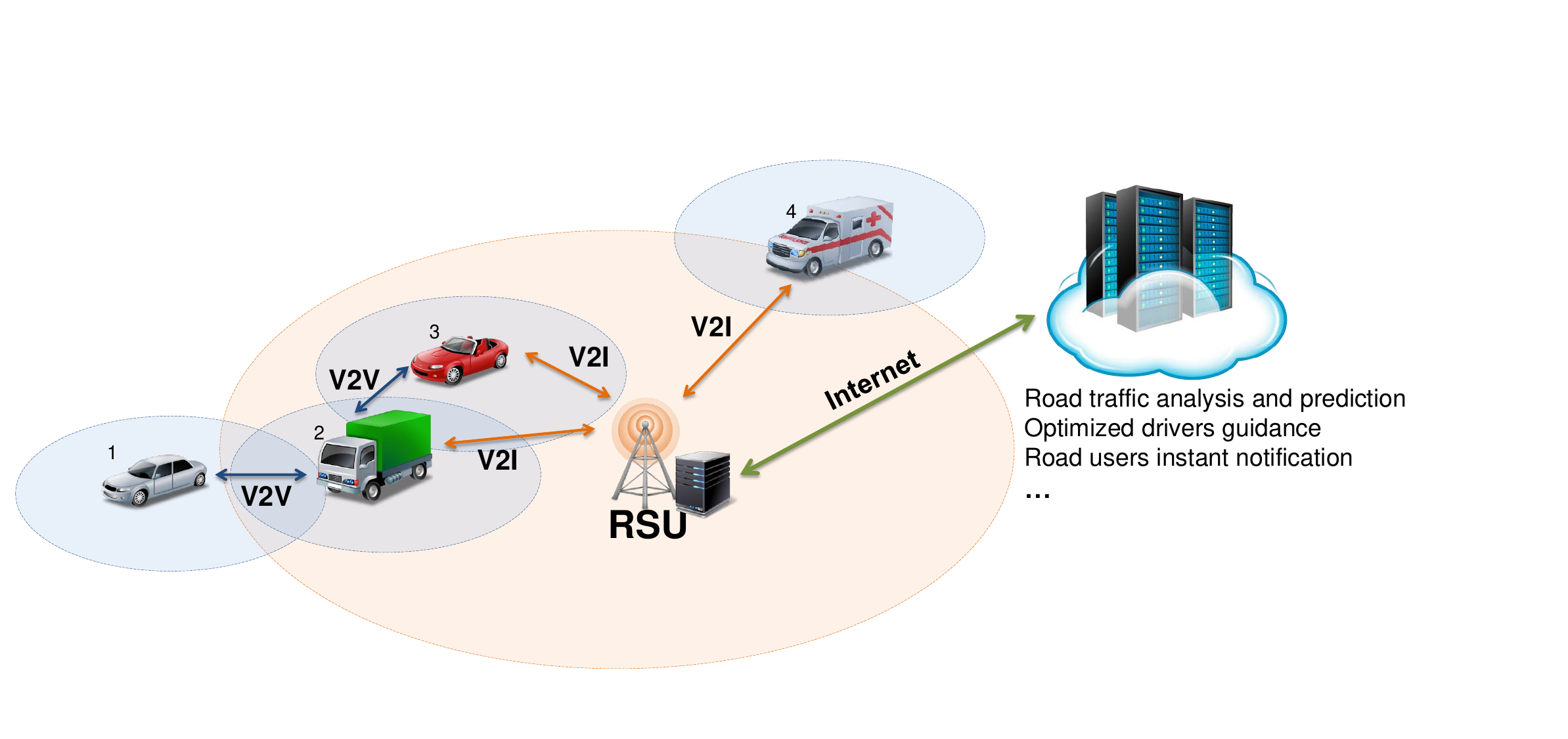}
\caption{Global VANET architecture.}
\label{fig:vanet-architecture}
\end{figure}

Figure~\ref{fig:vanet-architecture} illustrates a typical VANET scenario and the importance of including RSUs in the VANET architecture. 
In the figure, the coverage of the OBUs is shaded in blue, and represents the maximum distance in which two vehicles may utilize V2V to communicate with each other (i.e., vehicles 2 and 3 and vehicles 1 and 2 are the only ones that are able to exchange information with each other). 
The communication range of the RSU is shaded in orange, and therefore, vehicles 2, 3, and 4 can communicate with the RSU via V2I. 
In the presented example, the ambulance (vehicle 4) is approaching to vehicles 1, 2, and~3. These vehicles are outside the coverage of the OBU of the ambulance, therefore V2V communications cannot be used. The only way to warn that the ambulance is approaching is by using a RSU to forward messages to vehicles 2 and 3. Thus, the RSU extends the effective communication range of the ambulance. 
In turn, the RSU can inform vehicles about possible new and more efficient routes, helping to improve the ambulance trip. 
Furthermore, all vehicles in the scenario can use the RSU connectivity to access traffic services, Internet, etc. 

Consequently, the deployment of a fixed infrastructure of RSUs along the roads is of vital importance when deploying modern and powerful ITS, which helps to mitigate the serious road traffic problems that have to be confronted in modern cities.

Deploying the RSU infrastructure for VANETs is a challenge because network designers must decide about the number, type, and location of RSUs to maximize quality-of-service (QoS) of the VANET, while satisfying and/or minimizing the deployment cost requirements. 
At this point, the network designers have to take into account that VANETs are used by different types of applications and services (safety and non-safety), and therefore, the final effective QoS of the network has to satisfy the requiered transmission data rates and delivery times of such applications. 

The RSU Deployment Problem (RSU-DP) consists in placing a set of RSU terminals in a given area. We study a multiobjective version of the RSU-DP, which proposes maximizing the network QoS and minimizing the deployment costs. This is a hard-to-solve optimization problem on city-scaled areas, as the number of possible solutions is very large\cite{Reis2014}. Heuristics and metaheuristics\cite{Nesmachnow2014}
are promising methods to deal with the RSU-DP; they allow computing good infrastructure designs in reduced execution times\cite{Cavalcante2012,Chengyuan2014}. 
In this article, we propose applying the NSGA-II evolutionary algorithm\cite{deb2002} to 
design the RSU infrastructure within a city-scaled network in M\'alaga (Spain). In order to obtain realistic results, we consider real information about road traffic (road map and traffic flow), hardware (network capabilities and costs), and VANET applications.

This article extends our previous conference paper\cite{Massobrio2015}, where the problem was first presented and preliminary results of applying a multiobjective evolutionary algorithm (MOEA) were reported.

The main contributions of the research reported in this article are: 
i) the multiobjective formulation of the RSU-DP considers in the QoS evaluation the maximum number of vehicles that can be simultaneously attended by a given RSU type, extending our previous study that just took into account the effective radio range of each RSU type;
ii) we solve realistic scenarios, larger than those previously solved in the related literature, and we include here the last new and updated traffic data published by the M\'alaga city council for 2015;
iii) we model a set of real VANET applications, considered in the QoS metric applied in the problem formulation to evaluate a set of potential locations for RSUs (these applications were not present in the previous conference paper); 
iv) we adapt two state-of-the-art heuristic methods (deterministic and randomized) for the problem, to be used as a baseline for the comparison of the proposed MOEA; 
v) we propose a parallel master-slave MOEA, including a new initialization operator based on the Randomized Knapsack algorithm as a novelty regarding our previous study; 
and vi) we report accurate results for cost and QoS for the problem instances solved:
the proposed NSGA-II is able to improve over the results computed by the best baseline heuristics up to 24.68\% and 52.71\% in terms of cost, and up to 34.09\% and 39.48\% in terms of QoS.

The article is organized as follows. Section~\ref{Sec:RSU-DP} introduces the multiobjective version of the RSU-DP. Section~\ref{Sec:RSU-RW} presents a review of works solving the RSU location problem and related radio network design problems. Section~\ref{Sec:MOEA} introduces the methods applied to solve the problem.
The specific features of the proposed MOEA to solve the RSU-DP are described in Section~\ref{Sec:Alg}. Section~\ref{Sec:Exp} describes the heuristic methods proposed as a baseline for comparing the results computed using the proposed MOEA, and reports the experimental evaluation of the proposed method on a set of realistic scenarios in the city of M\'alaga, using real infrastructure and VANET applications. Finally, Section~\ref{Sec:Conc} formulates the conclusions and the main lines for future work.

\vspace{-6pt}
\section{The RSU deployment problem}
\label{Sec:RSU-DP}
\vspace{-2pt}

The mathematical formulation of the RSU-DP considers the following elements:
\begin{itemize}
\itemsep1pt
\item A set of RSUs $R = \{R_1, \ldots, R_q\}$ to be installed in a city scenario for providing efficient VANET communications.
\item A set of RSU types $T = \{t_1,t_2,\ldots, t_l\}$.  Each RSU type is characterized by a given deployment cost and a coverage determined by the transmission power and the antenna gain. The type of a RSU is given by the function \emph{type}$: R \rightarrow T$.
\item A set of \textit{road segments} $S=\{s_1,s_2,\ldots,s_n\}$, which are potential locations for placing RSUs along the city streets. Each segment $s_i$ is defined by a pair of points $(p_j, p_k)$, with $p_j, p_k \in P=\{p_1,p_2,\ldots,p_m\}$. Each point $p_j$ is identified by its geographical coordinates (latitude, longitude). The length of a given segment $s_i$ is given by the function \emph{len}$: S \rightarrow \mathbb{R}^+$. RSUs can be placed at any location within each segment $s_i$.
\item An estimation of the number of vehicles per time period across each segment $s_i$, given by function $NV$$: S \rightarrow \mathbb{N}^+$, and the average vehicle speed for each segment, given by function \emph{sp}$: S \rightarrow \mathbb{R}^+$.
\item A cost function $C$$: T \rightarrow \mathbb{R}^+$, where $C(t_g)$ indicates the monetary cost of placing a RSU of type $t_g$ in the deployed infrastructure.
\item A set of applications $A=\{A_1,A_2,\ldots,A_u\}$ to be used over the VANET. Each application has specific QoS requirements, given by function $Q$$: A \rightarrow \mathbb{N}^{+}$$\times$ $\mathbb{N}^{+}$. $Q(A_h)$ is a vector with two elements, indicating the QoS requirements for \textit{packet delivery ratio} (PDR) and \textit{end-to-end delay} (E2ED) for application $A_h$. On a given scenario, $Q(A_h)$ is used to define the \textit{maximum number of users} to be served by each RSU, given by function \mbox{$MU$$: R \times A  \rightarrow \mathbb{N}^+$}.
\end{itemize}

Solutions of the problem are defined by a set of RSUs placed over the road segments of the city, represented by a set $ sol = \{R_1,R_2,\ldots, R_l\}$, where $l$ is the number of RSUs (\#\textit{RSU}) in solution $sol$ ($l \le n$). Each RSU is installed in a specific coordinate within a segment $s_i$. The segments covered by a RSU are given by the function \emph{cov}$: R \rightarrow S$, and the portion of segment $s_k$ covered by RSU $R_j$ is given by the function \emph{cp}$: R \times S \rightarrow [0,1]$.

The multiobjective version of the RSU-DP proposes to find a set of locations and the type of RSU to deploy in each location, with the goal of maximizing the \textit{service time} given by the whole RSU infrastructure, while simultaneously minimizing the \textit{total cost} of deployment. 
The service time is a metric related to the QoS offered to the VANET users. It is related to the number of vehicles attended by RSUs, the time they are served (considering the coverage and average speed per each road segment), and the type of applications used in the studied scenario.  

Formally, the problem is defined as the simultaneous optimization of two objective functions: 
maximize the QoS, given by $f_1(sol,A_h)$ (Equation~\ref{Eq:f1}) and minimize the cost, given by $f_2(sol)$ (Equation~\ref{Eq:f2}).
The corresponding values for function $MU$ for each RSU and application type are computed by simulations (see Section~\ref{sec:problem-instances}).

\begin{align}
\small
\text{max} & ~f_1 (sol, A_h) = \sum_{R_j}^{R_j \in \text{ } sol} \max \left( MU(R_j,A_h), \sum_{s_i \in cov(R_j)} NV(s_i) \times 
\frac{cp(R_j,s_i)\times len(s_i)}{sp(s_i)} \right) \label{Eq:f1}
\end{align}
\vspace{-6pt}
\begin{align}
\small
\label{Eq:f2}
\text{min}  &~f_2 (sol) = \sum_{R_j}^{R_j \in \text{ } sol} C(type(R_j))
\end{align}

\vspace{-13pt}

\section{Related work}
\label{Sec:RSU-RW}

\vspace{-4pt}

Including RSUs in the network loop improves the global VANET performance in terms of connectivity, transmission delays, and communication ranges\cite{Liang2012}. Deploying a low cost and high coverage RSU infrastructure is often a capital issue for the success of VANETs in real cities.
This section reviews computational intelligence methods applied to the RSU-DP and related problems.

In the related literature, different studies address the RSU-DP.  Most of these works analyze RSU-DP as a version of the Radio Network Design problem\cite{Mendes2009}. 
However, as most nodes in VANETs are vehicles, the design of the roadside platform prioritizes locations taking into account road traffic information as speed of the vehicles, traffic density, etc.

Both exact methods and heuristics have been applied to solve the RSU-DP and related problems.

Trullols et al.\cite{Trullols2010} defined the Maximum Coverage with Time Threshold Problem (MCTTP) to maximize the number of vehicles that get in contact with a given number of RSUs for a given amount of time over a certain area. The authors proposed three greedy algorithms with different knowledge of the road topology and identity of the vehicles. These approaches were applied over a scenario with real road and mobility data from Zurich, Switzerland. The results showed that knowledge of vehicular mobility is the main factor to achieve an almost-optimal roadside deployment. Given such knowledge, the heuristics successfully planned a deployment capable of informing more than 95\% of vehicles.

Aslam et al.\cite{Aslam2012} applied the Balloon Expansion Heuristic (BEH) and Binary Integer Programming (BIP) 
to minimize the reporting time, installing a fixed number of RSUs in Miami, USA. The methods used information about speed, traffic density, and likelihood of incidents.
BEH performed better than BIP in the reported experiments.

A Voronoi-based algorithm was applied by Patil and Gokhale\cite{Patil2013} to optimize packet loss, communications delays, and network coverage, while minimizing the number of RSUs required in a deployed vehicular network in an area of Nashville, USA.    
The algorithm used information about the speed of vehicles and the traffic density to evaluate the solutions. 

Ben Brahim et al.\cite{Brahim2014} solved a variant of the RSU-DP in Doha, Qatar, considering the traffic network as a graph with weighted links. The weight of the links are computed according to road traffic and mobility-based parameters, such as road traffic density and average speed. Afterwards, all potential positions for the RSUs are computed by applying two different approaches: a dynamic algorithm based on 0-1 Knapsack problem (KP\_DynAlg) solver and the PageRank algorithm. The KP\_DynAlg improved over the results computed by the PageRank method.  

Some studies have proposed applying evolutionary algorithms (EAs) for solving variants of the RSU-DP, in order to obtain accurate solutions while consuming reasonable computational resources. An early approach studied applying a genetic algorithm (GA) that uses a VANET simulator to evaluate the QoS of the computed solutions in a given area of 16$\times$16 km$^2$ in the city of Brunswick, Germany, with about 500 km of roads and 10000 vehicles\cite{Lochert2008}. 
The authors introduce a domain aggregation scheme to minimize the required overall bandwidth for the VANET and propose a GA to locate static roadside units (called \textit{supporting units}) to deal with a highly partitioned VANET in an early deployment stage. The proposed GA was useful to improve the travel time savings achieved by a given vector of active SU locations.

Cavalcante et al.\cite{Cavalcante2012} compared GA against the greedy approach proposed by Trullols et al.\cite{Trullols2010} to solve the MCTTP, taking into account real data form four different regions: Zurich downtown, Winterthur, Baden, and Baar. The proposed GA uses a greedy method to initialize the population. The results showed that the GA solutions obtained better vehicle coverage: up to 11\% better than those computed by the greedy approach.

Another GA proposal is by Cheng et al.\cite{Cheng2013}, who used geometry-based coverage information about the roads (without vehicles mobility related data) of Yukon Territory, Canada, for the solution evaluation. 
The GA computed the fitness in terms of the ratio between the covered road area and the whole road area, computed using a square grid of $1m \times 1m$. 
This approach improved the results computed by the $\alpha$\emph{--coverage} algorithm, which proposes placing the RSUs in the center of the junctions.

A summary of the main related works about heuristics and computational intelligence methods (following evolutionary and non-evolutionary approaches) to solve the RSU-DP and related problems is presented in Table~\ref{Tab:RW}.

\begin{table}[!h]
\caption{Summary: related work about heuristics and computational intelligence methods applied to the RSU-DP.}
\centering
{\small
\setlength{\extrarowheight}{2pt}
\setlength\tabcolsep{3pt}
\begin{tabular}{lrlll}
\toprule
\textit{author} & \textit{year} & \textit{problem} & \textit{method} & \textit{scenario} \\
\hline
\multicolumn{5}{c}{non-evolutionary approaches}\\
\hline
Trullols et al.\cite{Trullols2010} & 2010 & vehicle maximization & greedy algorithms & Zurich, Switzerland\\
Aslam et al.\cite{Aslam2012} & 2012 & RSU installation & BEH, BIP & Miami, USA\\
Patil and Gokhale\cite{Patil2013} & 2013 & RSU optimization & Voronoi-based algorithm & Nashville, USA \\
Ben Brahim et al.\cite{Brahim2014} & 2014 & mobility/traffic based & Knapsack, PageRank & Doha, Qatar \\
\hline
\multicolumn{5}{c}{evolutionary approaches}\\
\hline
Lochert et al.\cite{Lochert2008}& 2008 & supporting units location & GA & Brunswick, Germany \\
Cavalcante et al.\cite{Cavalcante2012} & 2012 & MCTTP--RSU-DP & GA & Switzerland cities\\
Cheng et al.\cite{Cheng2013} & 2013 & geometry-based coverage & GA & Yukon territory \\
Massobrio et al.\cite{Massobrio2015b,Massobrio2015} & 2015 & multiobjective RSU-DP & MOEA & M\'alaga, Spain \\
\bottomrule
\end{tabular}}
\label{Tab:RW}
\end{table}

Our previous works\cite{Massobrio2015b,Massobrio2015} were the first studies that applied an explicit multiobjective approach to solve the RSU-DP. Our proposal was oriented to maximize the coverage, in terms of the time that vehicles are connected to the RSUs, and minimize the deployment cost. We consider real information concerning both traffic (speed, traffic density, and road map) and hardware (costs and capabilities) for the case of urban locations in M\'alaga, Spain. The proposed MOEA obtained significantly better results than ad hoc greedy approaches, but the computed solutions did not cover the map properly, focusing on streets with high number of vehicles instead.

In this article, we extend our previous work\cite{Massobrio2015} by considering a more realistic QoS model that includes a set of different VANET applications, which are taken into account to compute the maximum number of vehicles that can be simultaneously attended by a given RSU type.
A more comprehensive experimental analysis is performed, including updated traffic data and modeling real VANET applications. Finally, the results are compared against those computed using specific heuristics, adapted from the work of Ben Brahim et al.\cite{Brahim2014}, in terms of cost, QoS, and multiobjective optimization metrics.

\vspace{-6pt}
\section{Metaheuristics and evolutionary computation}
\label{Sec:MOEA}
\vspace{-2pt}

This section introduces the methods applied to solve the problem: metaheuristics, evolutionary algorithms and multiobjective evolutionary algorithms.

\vspace{-0.1cm}
\subsection{Metaheuristics}
\vspace{-0.1cm}

Metaheuristics are strategies to define algorithmic frameworks that allow designing efficient 
techniques to find approximate solutions for search, optimization, and learning problems\cite{Glover1986}. They define high-level, heuristic-based, soft computing methods that can be applied to solve different optimization problems, by instantiating a generic resolution procedure\cite{Nesmachnow2014}.

In practice, many optimization problems arising in nowadays real-world applications from science and technology are NP-hard and intrinsically complex. A lot of computing effort is demanded to solve them, due to a number of reasons:  they have very large-dimension search spaces, they include hard constraints that make the search space very sparse, they are multimodal or multiobjective problems taking into account hard-to-evaluate optimization functions, or they manage very large volumes of data. This is the case for the problem solved in this article: the deployment of roadside infrastructure for VANETs, which is a variant of the well-known Radio Network Design problem\cite{Mendes2009}.

Metaheuristics provide efficient and accurate methods for solving realistic instances of the problem, that often cannot be solved 
using exact optimization methods
that are extremely time-consuming. In this article, we apply a multiobjective evolutionary metaheuristic to solve the RSU-DP. The main features of EAs and their multiobjective variants are described next.

\vspace{-0.1cm}
\subsection{Evolutionary algorithms}
\vspace{-0.1cm}

EAs are non-deterministic methods that emulate the evolutionary process of species in nature to solve optimization, search, and other related problems\cite{BFM97,Goldberg89}. In the last thirty years, EAs have been successfully applied for solving problems underlying many real and complex applications.

Algorithm 1 shows the generic schema of an EA. It is an iterative technique (each iteration is called a \emph{generation}) that works by applying stochastic operators on a set of  \textit{individuals} (the population \emph{P}) in order to improve their {\em fitness}, a measure 
that evaluates how good is a solution to solve the problem. Every individual in the population encodes a candidate solution for the problem. The initial population is generated by a random method or by using a specific heuristic for the problem (line 2 in Algorithm 1). An evaluation function associates a fitness value to every individual
(line 4). 
The search is guided by a probabilistic selection-of-the-best technique (for both parents and offspring) to tentative solutions of higher quality (line 5).
Iteratively, solutions are modified by the probabilistic application of \emph{variation operators} (line 6), 
including the \textit{recombination} of parts from two individuals or random changes (\textit{mutations}) in their contents, which are applied for building new solutions during the search.

The stopping criterion usually involves a fixed number of generations or execution time, a quality threshold on the best fitness value, or the detection of a stagnation situation. Specific policies are used to select the groups of individuals to recombine (the \emph{selection} method) and to determine which new individuals are inserted in the population in each new generation (the \emph{replacement} criterion). The EA returns the best solution ever found in the iterative process, taking into account the fitness function.

\begin{algorithm}
\caption{Generic schema for an EA.}\label{EAalg}
\begin{algorithmic}[1]
\State $t$ $\leftarrow$ 0 \Comment{Generation counter}
\State \textbf{initialize}($P$(0))
\While{not {stopcriterion}}
\State {\bf evaluate}($P$($t$))
\State {parents} $\leftarrow$ {\bf selection}($P$($t$))
\State {offspring} $\leftarrow$ {\bf variation operators}({parents})
\State $P$($t$+1) $\leftarrow$ {\bf replacement}({offspring}, $P$($t$))
\State $t$  $\leftarrow$  $t$ + 1
\EndWhile
\State {\bf return} best solution ever found
\end{algorithmic}
\end{algorithm}

One of the most popular variants of EA in the literature is the genetic algorithm (GA), which has been extensively used to solve optimization problems mainly due to its simplicity and versatility.

The classic GA formulation was presented by Goldberg\cite{Goldberg89}. Based on the generic schema of an EA, 
a GA defines selection, recombination and mutation operators, applying them to the population of potential solutions in each generation. In a classic application of a GA, the recombination operator is mainly used to guide the search (by exploiting the characteristics of suitable individuals), while the mutation is used as the operator aimed at providing diversity for exploring different zones of the search space. 

Parallel models for metaheuristics and EAs have been proposed to speed up the computing time required for the search when dealing with complex objective functions and hard search spaces\cite{Alba2013}. In this work, we apply a  master-slave model for parallelization, in order to reduce the execution time of computing the QoS objective when solving the RSU-DP problem (see details in Section~\ref{Sec:Alg}).

\subsection{Multiobjective evolutionary algorithms and NSGA-II}

\sloppy Multiobjective evolutionary algorithms \mbox{(MOEAs)\cite{Coello2002,deb2002}} are specific evolutionary optimization methods conceived to solve problems with many conflicting objective functions. MOEAs have obtained accurate results when used to solve difficult real-life optimization problems in many research areas.  

Unlike many traditional methods for multiobjective optimization,  MOEAs  find a set  with several solutions  in a single execution, since they work with a population of tentative solutions. 
MOEAs are designed 
to fulfill
two goals at the same time: 
\textit{i}) approximate the Pareto front, and \textit{ii}) maintain diversity, instead of converging to a section of the Pareto front. A Pareto-based evolutionary search leads to the first goal, while the second is accomplished using specific techniques 
from
multi-modal function optimization (e.g., sharing, crowding, etc.).

In this work, we apply NSGA-II (Non-dominated Sorting Genetic Algorithm, version II)\cite{Deb2001}, a popular state-of-the-art MOEA that has been successfully applied in many areas. A schema of NSGA-II is presented in Algorithm~2 (where $N$ is the population size). The fitness calculation is based on Pareto dominance, building \textit{fronts} of solutions. The evolutionary search on NSGA-II improves over the previous version (NSGA), using:
\textit{i}) a non-dominated, elitist sorting that reduces the complexity of the dominance check; \textit{ii}) a crowding technique for diversity preservation; and \textit{iii}) a fitness assignment that considers crowding distance values. 

\begin{algorithm}
\caption{Schema of the NSGA-II algorithm.}\label{MOEAalg}
\begin{algorithmic}[1]
\State $t$ $\leftarrow$ 0 \Comment{Generation counter}
\State offspring $\leftarrow$ $\emptyset$
\State {\bf initialize}($P$(0))
\While{not {stopcriterion}}
\State {\bf evaluate}($P$($t$))
\State R $\leftarrow$ $P$($t$) $\cup$ offspring
\State fronts $\leftarrow$ {\bf non-dominated sorting}(R))
\State $P$($t$+1) $\leftarrow$ $\emptyset$

\State $i$ $\leftarrow$ 1

\While{$\lvert P(t+1)\rvert + \lvert fronts(i)\rvert \leq N$}
\State {\bf crowding distance} (fronts(i))
\State $P$($t$+1) $\leftarrow$ $P$($t$+1) $\cup$ fronts(i)
\State $i$ $\leftarrow$ $i$+1
\EndWhile

\State {\bf sorting by distance} (fronts(i))
\State $P$($t$+1) $\leftarrow$ $P$($t$+1) $\cup$ fronts(i)[1:(N - $\lvert$$P$($t$+1)$\rvert$)]
\State selected $\leftarrow$ \textbf{selection}($P$($t$+1))
\State offspring $\leftarrow$ \textbf{variation operators}(selected)
\State $t$  $\leftarrow$  $t$ + 1
\EndWhile
\State {\bf return} computed Pareto front
\end{algorithmic}
\end{algorithm}

The NSGA-II algorithm proposed in this work has been engineered to compute accurate solutions for the RSU-DP. The main implementation details are presented in the next section.

\vspace{-6pt}
\section{The proposed NSGA-II algorithm for the RSU-DP}
\label{Sec:Alg}
\vspace{-2pt}

This section presents the details of the proposed NSGA-II evolutionary algorithm for the RSU-DP.

\subsection{Solution encoding}
\label{sec_sol_encoding}
In the proposed NSGA-II, solutions are represented as vectors of real numbers, having length $n = \#S$ (the number of elements in the set of road segments $S$). Each position on the vector contains the information about the RSU to install (if any) on the corresponding segment: i) the type of the RSU is given by the integer part of the real number (0 stands for the absence of RSU in the considered segment, and integers $1 \ldots k$ represent types $t_1 \ldots t_k$, respectively); and ii) the condidate location to install the RSU within the segment  is given by the fractional part of the real number, mapping the interval $[0,1)$ to points in the segment $[p_j, p_i)$.

\begin{figure}[!ht]
\centering
\includegraphics[width=0.9\linewidth]{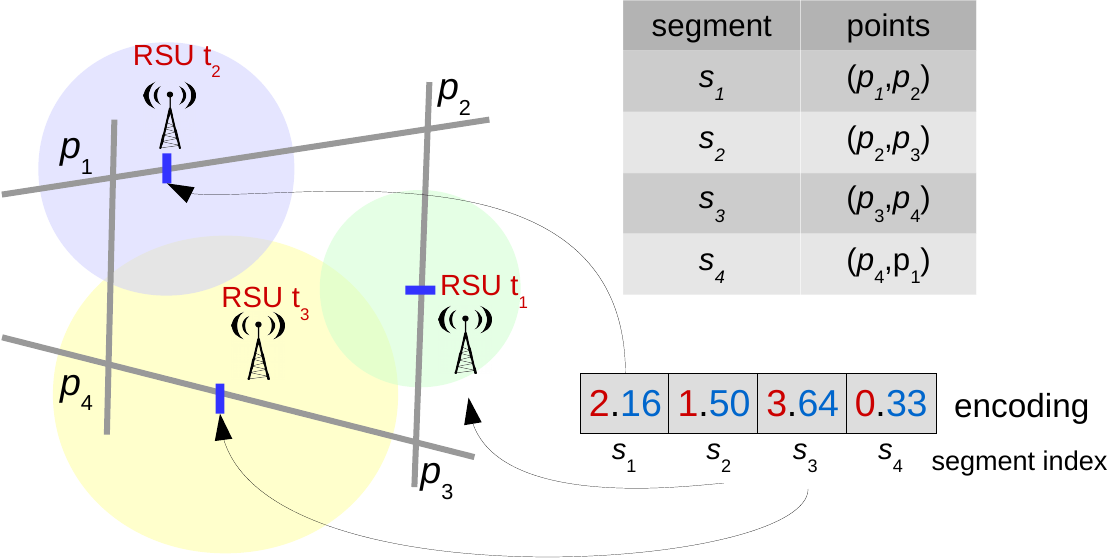}
\caption{Encoding for RSU-DP solutions.}
\label{Fig:encoding}
\end{figure}

Figure~\ref{Fig:encoding} shows an example encoding for a scenario consisting of four segments, where three RSUs are placed. For instance, the value 1.50 in position 2 of the vector indicate that the solution proposes to install a RSU of type 1 (integer part of 1.50) at the middle (fractional part of 1.50 = 0.50) of segment $s_2=(p_2,p_3)$.
The same holds for value 2.16 in the first position of the vector, corresponding to segment $s_1$, where a RSU of type 2 is installed at $0.16$$\times$$len(s_1)$ within segment $s_1=(p_1,p_2)$. Finally, the value 0.33 in the fourth position of the vector indicates that the solution proposes not installing a RSU in segment $s_4=(p_1,p_4)$ (the fractional part of the value encoded is irrelevant in this case). In fact, considering the covering radios of the RSUs placed in the map in Figure~\ref{Fig:encoding}, we see that it could be a wise decision, because segment $s_4$ is fully covered by RSUs $t_2$ and $t_3$ (of course, this decision reduces the installation cost, but the QoS for users might be reduced, depending on the number of vehicles and the applications used in the considered scenario).
  
\subsection{Evolutionary operators} 

The proposed MOEA applies different evolutionary operators and a parallel model to efficiently address the RSU-DP problem. This section describes such operators and parallel model. 

\subsubsection{Initialization} Instead of starting from a 
random set of solutions, we decided to seed the initial population
with the solutions computed by 
the Randomized Knapsack heuristic, explained in Section~\ref{SubSec:Heuristics}. 
This decision allows focusing the evolutionary search on a subspace of good quality solutions. The Randomized Knapsack is a constructive method, providing a range of partial solutions that are useful to generate the initial population of the MOEA. Particularly, we employ solutions with a high number of RSUs, since we are not interested in exploring the areas of the Pareto front with negligible values for QoS (this region has configurations that are not useful in practice, in a real-world scenario).

\subsubsection{Selection} The selection operator used is the tournament selection, as originally proposed in the NSGA-II algorithm\cite{deb2002}. The tournament size is two individuals and the fittest individual survives.

\subsubsection{Exploitation: recombination} The recombination operator used is the well-known two-point crossover (2PX), where offspring are generated by swapping genes in the parents' chromosomes that fall between two randomly selected cutting points.

\subsubsection{Exploration: mutation} 
\begin{figure}[!htbp]
\centering
\begin{subfigure}{\textwidth}
\centering
\includegraphics[width=0.7\linewidth]{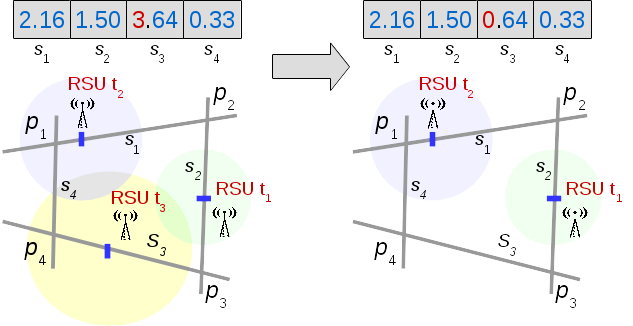}
\caption{Mutation applied with probability $\pi_A$ (remove RSU).}
\label{Fig:mutation1}
\end{subfigure}
\vspace{10pt}

\begin{subfigure}{\textwidth}
\centering
\includegraphics[width=0.7\linewidth]{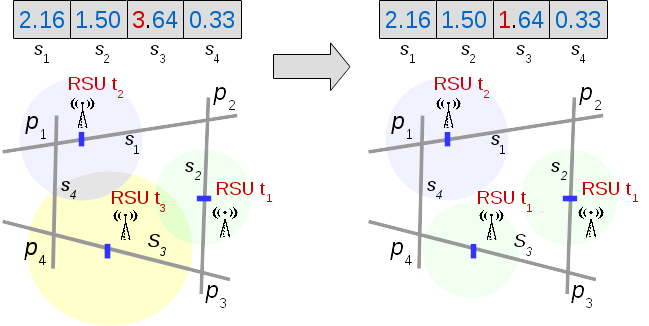}
\caption{Mutation applied with probability $\pi_B$ (modify RSU type).}
\label{Fig:mutation2}
\end{subfigure}
\vspace{10pt}

\begin{subfigure}{\textwidth}
\centering
\includegraphics[width=0.7\linewidth]{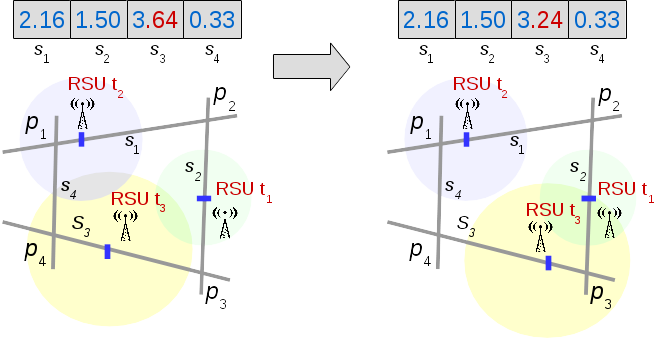}
\caption{Mutation applied with probability $\pi_C$ (modify RSU location within segment).}
\label{Fig:mutation3}
\end{subfigure}
\caption{Variations applied by the mutation operator. }
\label{Fig:mutation_all}
\end{figure}

We designed an ad-hoc mutation operator in order to provide enough diversity to the search, avoiding NSGA-II to get stuck in a specific region of the Pareto front. The mutation operator probabilistically applies three variations on solutions. This variations work as follows: 
\begin{enumerate}
\item With probability $\pi_A$, the mutation operator changes the integer part of the selected gene value to 0, thus removing the RSU (if any) from the corresponding segment (see Figure~\ref{Fig:mutation1}). 
\item With probability $\pi_B$, the mutation operator changes the integer part of the selected gene value for a different one randomly picked in $[1,k]$, thus changing the type of the RSU (or adding one if there was none) to a random type picked uniformly in $T$ (see Figure~\ref{Fig:mutation2}). 
\item With probability $1-\pi_A-\pi_B$, the variation applied corresponds to a Gaussian Mutation with a standard deviation of $\sigma$ to the selected gene value, thus changing the position of the RSU within the segment (see Figure~\ref{Fig:mutation3}).
\end{enumerate}

\subsubsection{Parallel model} 
We apply a master-slave parallel model for metaheuristics\cite{Alba2013} to reduce the execution time demanded to evaluate the objective functions for each individual in the population. The decomposition approach allows NSGA-II to efficiently compute the objective functions for the set of candidate solutions in the population, as described in the next subsection.

\subsection{Computing the objective functions}

In order to compute the two objective functions to be optimized in RSU-DP, two functions were evaluated: the installation cost and the quality of service. These two functions are presented next.

\subsubsection{Installation cost}
The total installation cost is simply computed by adding the cost of each RSU placed in the solution, taking into account the corresponding RSU type. 

\subsubsection{Quality of service} 
\begin{figure}[!h]
\vspace{-2pt}
\centering
\includegraphics[width=0.48\linewidth]{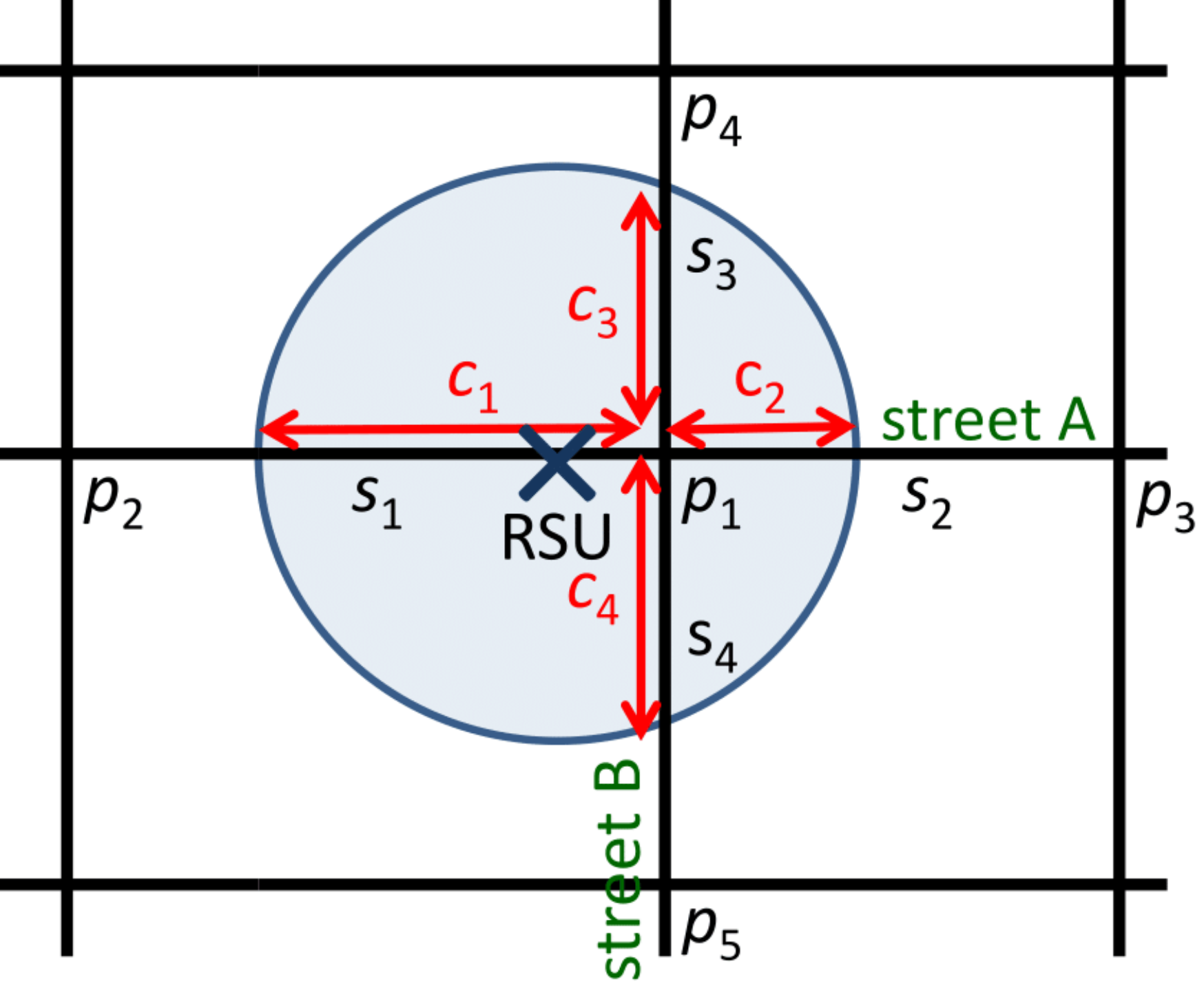}
\vspace{-1pt}
  \caption{Calculation of the vehicles attended by a RSU.}
  \label{Fig:coverage}
\vspace{-2pt}
\end{figure}

For computing the QoS, we consider the distances and values shown in the diagram in Figure~\ref{Fig:coverage} %(intersection of street A and street B). 
(intersection of streets A and B). 
The RSU placed in the point ``$\times$" covers the subsegments $c_1$ (in $s_1$), $c_2$ (in $s_2$), both in street A, and $c_3$ (in $s_3$), and $c_4$ (in $s_4)$, both in street B. The number of effective vehicles attended is computed by 
$\sum_{i=1}^{i=4} NV(s_i) \times \frac{c_i}{sp(s_i) }$. 
The computation requires finding the intersections between the road segments and the circle defining the coverage of the RSU. Given that the distances involved 
are relatively small, we use straight lines in the latitude-longitude space as an estimation, with negligible error. This approximation makes the computation faster, improving the overall performance of the algorithm. 

In order to avoid situations where the fitness function computations take into account several times 
the same area (i.e., some vehicles in that area are counted multiple times), it is necessary to keep track of the number of vehicles that each installed RSU has already served. 
That number of vehicles depends on the type of the data streams that the RSU is providing (data, voice or video), since the minimum QoS required by each type of stream limits the maximum number of vehicles that can be simultaneously attended by a given RSU type.

\vspace{-6pt}
\section{Experimental analysis}
\label{Sec:Exp}
\vspace{-2pt}

This section presents the details of the experimental analysis performed to evaluate the proposed NSGA-II to solve the RSU-DP.

\vspace{-2pt}
\subsection{Development and execution platform}
\vspace{-2pt}

The proposed MOEA was implemented using the ECJ library, a Java-based evolutionary computation research system developed at ECLab Evolutionary Computation Laboratory, George Mason University\cite{White2011}.
ECJ includes easily modifiable classes for solving multiobjective optimization problems using the NSGA-II algorithm.

The experimental analysis was performed over an AMD Opteron 6172 2.10 GHz server, with  24 cores and 24 GB RAM at Cluster FING, the High Performance Computing facility at Universidad de la Rep\'ublica, Uruguay\cite{Nesmachnow2010}. 

Since computing the fitness of an individual is highly CPU-intensive, the evaluation of the population is performed in parallel using 24 Java-threads. Thus, each thread evaluates 3 individuals of the population. 

For each probem instance (map, traffic, and application), we performed 30~independent executions of the proposed MOEA. For the heuristics used as a baseline for the results comparison, we performed one execution for the PageRank algorithm (deterministic) and 30 independent executions for the Knapsack algorithm (randomized). Including the parameter setting and validation experiments, we performed a total number of \textbf{1080} executions of NSGA-II and \textbf{279} executions of the heuristic algorithms.

\subsection{Problem instances}
\label{sec:problem-instances}

In order to evaluate the proposed MOEA, we defined a real world problem scenario that is relevant for our community. We included real information for a number of elements: a real map of the city of M\'alaga, real road traffic data from the M\'alaga Council, real RSU network interfaces/antennas, and real applications to be executed over the VANET. The main details about these elements are presented next.

\subsubsection{Map} 
\begin{figure}[!h]
\centering
\includegraphics[width=0.97\linewidth]{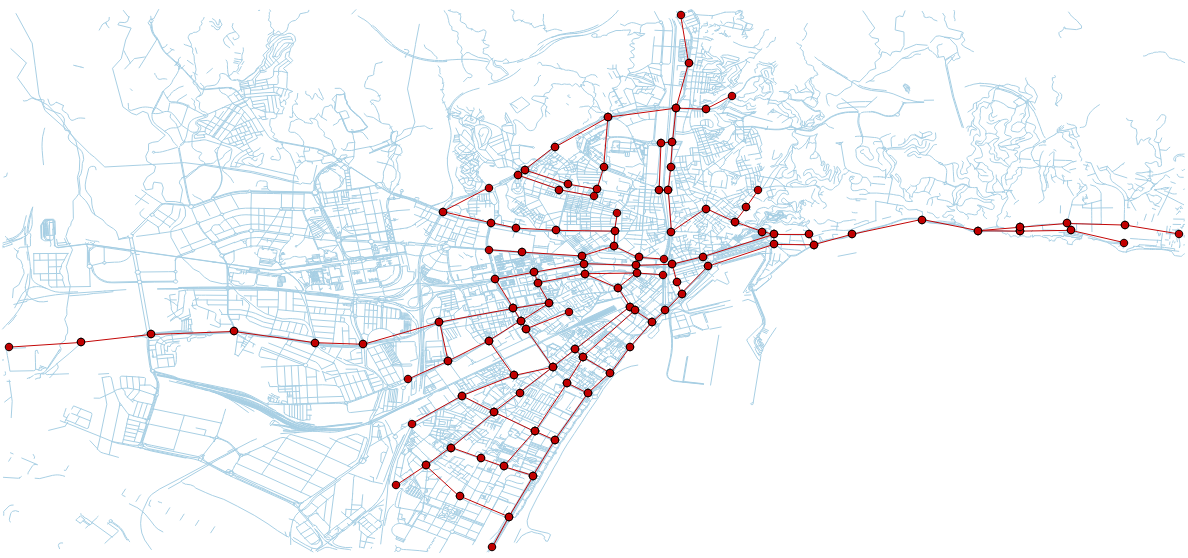}
  \caption{Segments defined over the real map of M\'alaga.}
  \label{Fig:map}
\end{figure}

Figure~\ref{Fig:map} shows the map of M\'alaga considered in the experimental analysis. The map covers an area of 42557 km$^2$ in the city, including a total number of 106 points, which define 128 segments with lengths between 55.5 and 1248.2~m, and an average length of 483.9~m. All major traffic ways, including avenues and important streets in M\'alaga are sampled. Some important avenues with large traffic volume define multiple segments in the map (e.g., \textit{Avenida de Andaluc\'ia}, \textit{Avenida de Vel\'azquez}, \textit{Avenida de Valle Incl\'an} and \textit{Paseo Mar\'itimo Pablo Ruiz Picasso}, all of them with more than six segments defined in the map).

\subsubsection{Road traffic data}
The traffic data used in our experiments are based on the information collected by the M\'alaga City Council using a set of sensors located along the roads. These sensors returned the total number of vehicles that circulated during the first six months of 2015. The information is publicly available at the M\'alaga Council Mobility website\cite{webmovilidad}.

We used the traffic information to define the \textit{normal} traffic pattern in our RSU-DP scenario.
In addition, we applied two probabilistic multiplicative factors over the \textit{normal} pattern, to define a \textit{low} pattern, reducing the traffic randomly in [0--20\%] and a \textit{high} pattern, increasing the traffic randomly in [0--20\%]. These patterns represent situations with low and high road traffic density, respectively, according to the real data from the M\'alaga City Council (in fact, studying the traffic statistics for peak hours provided by the Council, we verified that that the number of vehicles for all main roads are about 20\% higher than in a normal traffic scenario).

\subsubsection{RSUs} 
The initial effort to standardize the DSRC radio technology took place in the ASTM~2313 working group in the U.S. This effort migrated to the IEEE 802.11 standard group that proposed IEEE 802.11p, which stands for \textit{wireless access in vehicular environments} (WAVE)\cite{Jiang2008}. Following these specifications, the RSUs considered in our study are equipped with a real world IEEE 802.11p commercial network interface. This hardware configures a realistic scenario and allows computing useful results from the point of view of both the research and technological communities. 

Each network interface is connected to an external antenna to improve the communication capabilities, according to a given antenna gain. The gain, measured in decibels (dBi), is a measure of the power of the radio signal radiating from the antenna. Generally, the higher the gain of an antenna, the longer the radio range that can be obtained, and the better the QoS provided by the infrastructure.

The RSUs analyzed in the problem instances solved in the experimental analysis differ in the antenna connected. Three different commercial omni-directional antennas, which can be found in online shops (e.g., Cetacea\cite{cetacea}), are considered. The antennas differ in the gain offered and the cost. A summary of the main features of such antennas is presented in Table~\ref{tab:antennas}. 
Our study does not exclude the possibility of incoporating new communication devices (network interfaces and antennas) for the city infrastructure. The proposed algorithms do not depend on the type or features of the RSUs considered. 

\begin{table}[!h]
\vspace{-0.1cm}
\centering
\caption{General information about the used antennas to define different RSU.}
\label{tab:antennas}
\small
\begin{tabular}{llrr}
\toprule
\textit{type} & \textit{commercial model} & \textit{gain}& \textit{cost} \\
\midrule
$t_1$ & Echo Series Omni Site 6dBi & 6 dBi & \$121.70\\
$t_2$ & Echo Series Omni Site 9dBi & 9 dBi & \$139.20\\
$t_3$ & Echo Series Omni Site 12dBi & 12 dBi & \$227.50\\
\bottomrule
\end{tabular}
\vspace{-0.2cm}
\end{table}

One of the main features that defines a given RSU is its \emph{effective radio range} (ERR). 
EER indicates the farthest distance at which the RSU may exchange data packets with the vehicles, and it is a relevant metric to evaluate the performance of the communications provided by the VANET infrastructure. In fact, ERR is one of the components we evaluate in the fitness function to compute the QoS provided by each RSU deployed in the studied scenario. 

To determine the ERR metric for each studied RSU, we performed realistic VANET simulations evaluating the 
PDR
at different distances (from 0 to 650~m). The experiments were performed by using the ns-2 simulator\cite{ns-2} to evaluate the communications. The simulated VANET scenario is defined by a given RSU and 10 moving cars at 40~Km/h (11.11~m/s) that utilize IEEE 802.11p network devices in a urban area. 
During the simulations, the RSU sent continuous data streams at 256 Kbps to the vehicles. The probabilistic Nakagami radio propagation model\cite{Saunders1999} was used to represent the channel fading characteristics of urban scenarios.  Each scenario was simulated 15~times to obtain robust average PDR values.  
In order to ensure a realistic QoS that guarantees reliable communications, we defined the ERR of each RSU as the distance at which the average PDR is equal or higher than 66.67\% (i.e., less than one packet lost for each three packets transmitted).

\begin{figure}[!h]
\vspace{-0.3cm}
\centering
\includegraphics[width=0.62\linewidth,trim={160 0 230 0},clip]{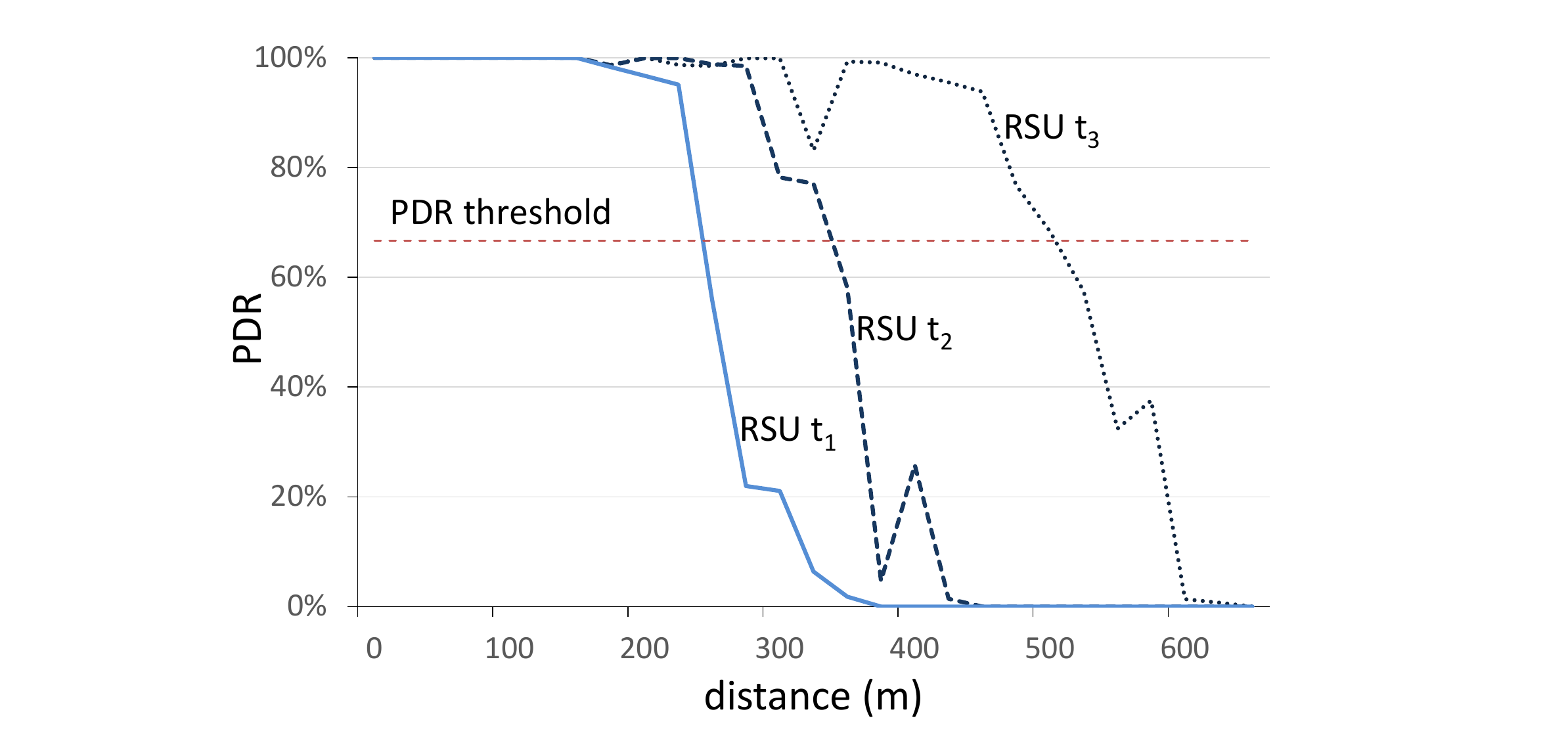}
\vspace{-1pt}
{\small
\setlength{\tabcolsep}{12pt}
\begin{tabularx}{0.55\textwidth}{lccc}
\midrule
\textit{type} & $t_1$ &  $t_2$ &  $t_3$ \\
\midrule 
\textit{ERR} & 243.12 m & 338.70 m & 503.93 m \\
\midrule
\end{tabularx}
}
\vspace{0.0cm}
\caption{ERR experimental results.}
\label{Fig:ERR-experiment}
\vspace{-0.2cm}
\end{figure}

Figure~\ref{Fig:ERR-experiment} reports the experimental results of the simulations to obtain the ERR for each RSU type. According to the PDR threshold defined, the ERR for the RSU type $t_1$ is 243.12 m, for RSU type $t_2$ is 338.70 m, and for RSU type $t_3$ is 503.93 m
(see the values at the bottom of Figure~\ref{Fig:ERR-experiment}).

\subsubsection{Applications}
As we introduced in Section~\ref{sec:intro}, VANET communications comprise mainly three types of applications: road safety, traffic efficiency, and infotainment. 
The two first types of applications rely on the exchange of small data packets with very short communication delays. 
The infotainment applications comprise a wide variety of applications that are principally based on receiving audio/voice and video streams\cite{Campolo2015,Hartenstein2010}. 

One of the main contributions of the study reported in this article is that we defined problem instances taking into account the requirements of the different types of VANET applications. This analysis allows the designers to prioritize the applications' constraints in the final RSU deployment. Thus, for each RSU type, the computed QoS metric takes into account the maximum number of vehicles that can be served while fulfilling the requirements for each VANET application type ($MU$). The QoS constraints for each VANET application type evaluated in this study are summarized in Table~\ref{tab:qos_requirements}, based on the study by Chantaksinopas et al.\cite{Chantaksinopas2012}

\begin{table}[!h]
\centering
\caption{QoS requirements of VANET applications taken into account in this study ($Q(A_i)$).}
\label{tab:qos_requirements}
\small
\setlength{\tabcolsep}{6pt}
\renewcommand{\arraystretch}{1.1}
\begin{tabular}{llll} 
\toprule
\textit{application type} & \textit{packet size (bytes)} & \multirow{1}{*}{\textit{generated data flow}}	& \multirow{1}{*}{\textit{QoS requirements}} \\
\midrule
data ($A_1$)	& \multirow{1}{*}{238 bytes}	&	\multirow{1}{*}{19 kbps (10 packets/s)}	&	E2ED$<$100 ms \& PDR=100\% \\

voice/audio  ($A_2$)	&	\multirow{1}{*}{238 bytes}	&	\multirow{1}{*}{25 kbps}	&	E2ED$<$400 ms \& PDR$>$16\% \\
video  ($A_3$) 	&  \multirow{1}{*}{791 bytes}		&	\multirow{1}{*}{384 kbps}	& E2ED$<$400 ms \& PDR$>$8.33\% \\
\bottomrule
\end{tabular}
\end{table}

In order to evaluate the $MU$ function described in Section~\ref{Sec:RSU-DP}, 
we defined an iterative procedure based on performing realistic urban VANET simulations for each RSU ($R_i$) and application ($A_j$). 

The procedure consists in simulating a given scenario that includes a RSU of a given type $R_i$ and  
different number of vehicles 
spread through a circular area with radio $ERR(R_i)$.
In these simulations, the VANET nodes generate data flows (traffic) according to the $A_i$ application (see the third column on Table~\ref{tab:qos_requirements}). 
Therefore, the VANET scenario is defined according three different parameters ($n$, $R_i$, $A_j$).  

The evaluation procedure starts by simulating the scenario with one vehicle (1, $R_i$, $A_j$) and computing the two relevant QoS metrics defined for the problem (PDR and E2ED). After that, a new vehicle is added and the new configuration is simulated. The iterative method stops when the computed QoS metrics do not accomplish the requirements defined in $Q(A_i)$ function. Finally, $MU(R_i, A_j)$ is the number of vehicles previous to the last one that was simulated. 
The computed results for each RSU and application type are summarized in Table~\ref{tab:qos_maxvehicles}.

\begin{table}[!h]
\centering
\caption{Number of vehicles that can be served for each RSU and application types.}
\label{tab:qos_maxvehicles}
\setlength{\tabcolsep}{8pt}
\begin{tabular}{lrrrr}
\toprule
	\multirow{2}{*}{\textit{RSU type}} & & \multicolumn{3}{c}{\textit{application type}}   \\ 
\cline{3-5}\\[-11pt]
	&	 & safety & voice/audio & video \\ 
\midrule	 
	$t_1$ & & 45 & 34 & 31 \\
	$t_2$ & & 45 & 44 & 34 \\
	$t_3$ & & 46 & 52 & 37 \\
\bottomrule
\end{tabular}
\end{table}

According to the results presented in Table~\ref{tab:qos_maxvehicles}, for example, all RSUs equipped with an antenna type $t_2$ can provide an adequate QoS to 34 vehicles executing a video stream application, and so on.

\subsection{Heuristic methods used as a baseline for the comparison}
\label{SubSec:Heuristics}

In order to evaluate the quality of the solutions computed by the proposed NSGA-II, we implemented two versions of well-known heuristic methods to solve a different variant of the RSU location problem, originally proposed by Ben Brahim et.~al\cite{Brahim2014}: the PageRank heuristic and the Knapsack algorithm. 

\subsubsection{Constructive PageRank heuristic}
PageRank is a voting algorithm, initially developed to compute the importance of web pages in the Internet by taking into account the number of inbound and outbound links from and to other web pages\cite{Langville2011}. 

The PageRank algorithm has been previously applied to solve the RSU deployment problem by Ben Brahim et al.\cite{Brahim2014}.  
In that study, the authors applied the PageRank version for weighted graphs to rank the potential locations for RSUs (road intersections) according to mobility-related information (e.g., traffic density, average speed of vehicles). The road traffic network is modeled as a graph with weighted links, which represent roads (segments), and vertices, which represent the intersections. The weight of each link is given by mobility-related information (e.g., density, average speed).

The \textit{weighted PageRank} is applied to a given directed graph $G=(V,E)$ defined by a set of vertices $V$, and a set of edges $E$,
The algorithm starts by setting the PageRank value of all vertices $v_i$ to a fixed value $d$: $PR^W(v_i)=d,\  \forall v_i \in V$. $d$ is known as the \textit{dumping parameter} and its default value is 0.85. Then, an iterative process is performed until a stop condition is reached (the convergence value is below a given threshold or a maximum number of iterations performed). 
In this iterative process, for a given vertex $v_i$, $PR^W(v_i)$ is computed by

\begin{equation}
\small
\label{eq:pagerank}
PR^W(v_i) = (1-d) + d \times \left( \sum_{v_j \in In(v_i)} w_{ij} \times \frac{PR^W(v_j)}{ \sum\limits_{v_k \in Out(v_j)} w_{jk}}\right)
\end{equation}
where $In(v_i)$ is the set of vertices that point to it (\textit{predecessors}), and $Out(v_i)$ is the set of vertices that $v_i$ points to (\textit{successors}), and $w_{ij}$ is the weight that for the edge that connects $v_i$ and $v_j$. 

In our study, we consider all the road segments as potential locations to install the RSUs, and not just the intersections as proposed by Ben Brahim et al.\cite{Brahim2014}. Therefore, we adapt the weighted PageRank algorithm with the purpose of sorting the segments (the edges of the graph) according to the rank value, and after that, applying a constructive heuristic over the sorted vector of segments. 

The weighted graph $G=(V,E)$ is defined by the set of points $P$ and the set of segments $S$  in the RSU-DP formulation ($V$=$P$ and $E$=$S$); 
The weight of each edge $w_{jk}$ is given by the weight of the represented segment $W(s_i)$, $s_i=(p_j,p_k)$, defined by Equation~\ref{eq:weighPR}.
\begin{equation}
%\small
\label{eq:weighPR}
w_{jk} = W(s_i) = NV(s_i) \times \frac{len(s_i)}{sp(s_i)}
\end{equation}
The rank value for each segment $s_i=(p_j,p_k)$ is 
computed as the sum of the PageRank values of $p_j$ and $p_k$, i.e., $SR^W(s_i) = PR^W(p_j) + PR^W(p_k)$.  
Thus, the segments are ranked in a sorted vector $S^{PR}$ in which $s^{PR}_i, s^{PR}_j \in S$, $i<j \Leftrightarrow SR^W(s^{PR}_i) > SR^W(s^{PR}_j$). 

Once the segments are sorted in $S^{PR}$, a constructive heuristic is applied to select and locate the RSUs. 
The heuristic iterates over the sorted vector $S^{PR}$ starting by the first segment ($s^{PR}_1$), which is the best ranked one. %by PageRank. 
For each segment $s^{PR}_i \in S^{PR}$, the different QoS provided for each one of the three RSU types when installing them in one of 10 different equidistant points within the segment that are located in positions $n \times 0.1 \times len(s^{PR}_i), n\in [0,9]$, are computed. Thus, the constructive algorithm evaluates the QoS of 30 different possible configurations (3 RSU types $\times$ 10 locations) in each segment.  It considers the configuration (RSU type and location) that provides the best QoS, if the whole QoS increases at least 1\% regarding the previous segment. Otherwise, the constructive PageRank heuristic does not locate any RSU in the current segment $s^{PR}_i$.

\subsubsection{Randomized Knapsack algorithm}
The 0-1 Knapsack problem\cite{Martello1990} is a well-known optimization problem
that
assumes having a bag with a capacity $W$ and a set of objects characterized by their benefit value $p_i$ and their wight $w_i$. 
The goal of the 0-1 Knapsack problem is to find the optimal subset of objects to include in the bag, maximizing the benefit $P$ without exceeding the capacity $W$.
  
The RSU-DP problem can be reduced to a 0-1 Knapsack problem, which can be solved applying a dynamic programming algorithm. 
In this work, we adapt the Knapsack algorithm by Ben Brahim et.~al\cite{Brahim2014} to solve the RSU-DP following a non-deterministic dynamic programming approach, the \textit{Randomized Knapsack} (RandKS) presented in Algorithm~3.

\begin{algorithm}
\caption{Schema of the RandKS($B$,SS,C,n,$Ksol$).}\label{RandKSalg}
\begin{algorithmic}[1]

\If{(n==0 or $B\leq$0)}
\State $Ksol$ $\leftarrow$ $\emptyset$ \Comment{No more segments or budget}
\State {\bf return} 0
\Else
\For{k$\leftarrow$1 \textbf{to} $K$} \Comment{For all RSU types}
\If{$B<$ $C(t_k)$} \Comment{No budget enough}
\State $cov_k$ = RandKS($B$,SS,C,n-1,$Ksol$) 
\Else 
\State $loc$ $\leftarrow random$ $\in$ [0,1)
\State $cov_k$ $\leftarrow$ coverage(($s_n$, $t_k$, loc),$Ksol$) + RandKS($B$-$C(t_k)$,SS,C,n-1,$Ksol$)  
\EndIf
\State rsu$\_$index $\leftarrow$ getIndex(max($cov_i$))
\If{{\bf not} rsu$\_$index==0}
\State $Ksol$ $\leftarrow$ $Ksol\ \cup$ ($s_n$, $t_k$, $location$)  
\EndIf
\EndFor
\EndIf
\end{algorithmic}
\end{algorithm}

\vspace{-2mm}
The Randomized Knapsack algorithm defines the set $SS$, which stores all the possible pairs of road segments and RSU types ($S \times R$), i.e., $SS = \{(s_1,t_1),(s_1,t_2),...,(s_1,t_K),...,(s_n,t_K)\}$. The elements of SS are the ones to include in the Knapsack bag when building a RSU infrastructure for a VANET.
The final solution is stored in the set $Ksol$, which stores tuples that include information about the installed RSUs: the road segment $s_i$, the RSU type $t_k$, and the $location$ of the RSU in the segment, i.e., $Ksol$ includes ($s_i$, $t_k$, $location$) tuples.

The \textit{location} \textit{loc} of the RSU in the segment is a real number in the range [0,1), that represents the relative value of the position in the segment, as explained for the solution encoding in NSGA-II in Section~\ref{sec_sol_encoding}.
In the original (deterministic) Knapsack algorithm by Ben Brahim et.~al\cite{Brahim2014}, the location of RSU is limited to the corners of the streets, i.e., the extremes of each segment. In our Randomized Knapsack algorithm, the location of the RSU is picked randomly within each segment, because we are working with a set of infinite possible locations, which cannot be explored one by one.

Two new functions are defined for the 
RandKS
algorithm: i) coverage(($s_n$, $t_k$, loc),$sol$), which computes the complete coverage provided by the RSUs stored in $sol$ plus the RSU located in the segment $s_n$ defined by 
($s_n$, $t_k$, loc); and ii) getIndex(max($cov_i$)), which returns the index of the RSU type that obtained the best (maximum) coverage.

\subsection{Multiobjective optimization metrics}
A large number of metrics have been proposed in the literature to evaluate MOEAs\cite{Coello2002,Deb2001}. 
In this work, we apply two relevant metrics in order to evaluate the results obtained by the NSGA-II algorithm:  hypervolume and relative hypervolume (RHV). These two metrics allows evaluating, in terms of both convergence and correct sampling, the set of non-dominated solutions of the problem.

The hypervolume measures the volume (in the objective functions space) covered by the computed Pareto front. 
The relative hypervolume is the ratio between the volumes (in the objective functions space) covered by the computed Pareto front 
and the true Pareto front. The ideal RHV value is 1.

The true Pareto front---which is unknown for the problem instances studied---is approximated by the set of non-dominated solutions found for each problem instance solved, in each one of the 30 execution of the proposed MOEA\cite{Coello2002,Deb2001}.

\subsection{Parametric configuration} 
Due to the stochastic nature of EAs, a parameter configuration is mandatory prior to the experimental analysis. For this purpose, we performed an experimental analysis to configure two important parameters of the NSGA--II algorithm: the crossover probability ($p_c$) and the mutation probability ($p_m$).

For the parameter setting, we use a set of problem instances which is different from the real-world problem instance, in order to avoid bias in the experimental analysis. We considered a smaller region of the map of M\'alaga, comprised of 74 segments, and we defined 3 different instances over this region corresponding to each type of application (data, voice, and video). For the parameter configuration we  considered the scenario with normal traffic. For each parameter, three candidate values were tested: $p_c \in \lbrace0.5, 0.7, 0.9\rbrace$, and $p_m \in \lbrace0.1, 0.01, 0.001\rbrace$.

We performed 30 independent executions of 10000 generations over each problem instance using each one of the different combinations of the  candidate parameter values, thus totalling 810 executions of the proposed MOEA.

Table~\ref{Tab:ConfParam} reports the hypervolume achieved using each parameter configuration on the scenario involving a video application, which is representative of the set of scenarios used in the parameter tuning.
The mean, median, and standard deviation ($\sigma$) are shown along with the minimum (\textit{min}) and maximum (\textit{max}) values achieved. Furthermore, the $p$-value corresponding to the Shapiro-Wilk test for normality is displayed as well as the \textit{Friedman rank} values corresponding to each parameter configuration.

The results from the Shapiro-Wilk test did not allow to confidently state whether the results samples follow a normal distribution (in six out of nine instances, the $p$-value was larger than 0.05). Therefore, the non-parametric Friedman rank test was used in order to compare the different parameter configurations against each other.
The $p$-values for the Friedman rank test did not allow to state whether there is one configuration that outperforms all the others with statistical significance. Therefore, we decided to use 
the configuration $p_C$ = 0.7; $p_M$ = 0.1, which
achieved the best mean and median hypervolume values.

\begin{table}[!h]
\caption{Parameter configuration results for the scenario with a video application.}
\centering
\setlength{\tabcolsep}{10pt}
\small
\begin{tabular}{cr|rrrrrrr}
\toprule
\multicolumn{ 1}{c}{\multirow{2}{*}{\textbf{$p_c$}}} & \multicolumn{ 1}{c|}{\multirow{2}{*}{\textbf{$p_m$}}} & \multicolumn{ 5}{c}{hypervolume ($\times 10^6$)} & \multicolumn{1}{c}{\textit{$p$-value}} & \multicolumn{1}{c}{\textit{Friedman}} \\
& &\multicolumn{1}{c}{\textit{mean}} & \multicolumn{1}{c}{\textit{median}} & \multicolumn{1}{c}{\textit{$\sigma$}} & \multicolumn{1}{c}{\textit{min}} & \multicolumn{1}{c}{\textit{max}} & \multicolumn{1}{c}{\textit{S-W}} & \multicolumn{1}{c}{\textit{Rank}}\\

\midrule
\multicolumn{ 1}{c}{\multirow{3}{*}{0.5}} & 0.001 & 9.27 & 9.27 & 0.04 & 9.18 & 9.36 & 0.99 & 148.00 \\
 & 0.01 & 9.27 & 9.28 & 0.05 & 9.16 & 9.36 & 0.86 & 160.00 \\
 & 0.1 & 9.26 & 9.27 & 0.05 & 9.11 & 9.34 & 0.04 & 147.00 \\ 
\midrule
\multicolumn{ 1}{c}{\multirow{3}{*}{\textbf{0.7}}} & 0.001 & 9.27 & 9.29 & 0.06 & 9.14 & 9.38 & 0.15 & 143.00 \\
 & 0.01 & 9.28 & 9.27 & 0.05 & 9.21 & 9.40 & 0.15 & 149.00 \\
 & \textbf{0.1} &  \textbf{9.29} & \textbf{9.30} & \textbf{0.05} & \textbf{9.20} & \textbf{9.38} & \textbf{0.76} & \textbf{171.00}\\ 
 \midrule
\multicolumn{ 1}{c}{\multirow{3}{*}{0.9}} & 0.001 & 9.25 & 9.26 & 0.07 & 9.07 & 9.36 & 0.34 & 134.00\\
 & 0.01 & 9.25 & 9.28 & 0.06 & 9.10 & 9.34 & 0.01 & 134.00 \\
 & 0.1 & 9.28 & 9.28 & 0.04 & 9.14 & 9.35 & 0.03 & 164.00 \\ 
\bottomrule
\end{tabular}
\label{Tab:ConfParam}
\end{table}

\vspace{-0.1cm}
\subsection{Numerical results}
\vspace{-0.1cm}

This subsection reports the numerical results achieved in the experimental evaluation of the proposed NSGA-II algorithm for the RSU-DP. The results shown are those corresponding to the 30 independent executions performed on each of the 9 problem instances studied: 3 different applications (data, voice, video), each with 3 different traffic levels (normal, high, low) for the proposed NSGA-II and the baseline heuristics (PageRank/Knapsack).

The RHV metric is used to compare the solutions computed by NSGA-II against the one computed sing the baseline heuristics. RHV is a good indicator of both convergence towards an ideal Pareto front and diversity among the set of non-dominated solutions.
The ideal Pareto front for a given problem instance is approximated with the set of non-dominated solutions computed by all algorithms on every independent execution of that instance.

In order to compare the RHV values achieved by each algorithm we used two statistical tests. 
We first performed the Shapiro-Wilk test to assess normality over the RHV values obtained by each algorithm on each problem instance.
In seven out of nine instances, the results from the Shapiro-Wilk test did not allow to state whether the samples follow a normal distribution or not. Therefore, the Friedman Rank test was used to assess whether the results achieved by one algorithm outperformed the others.

Table~\ref{Tab:RHV} reports the RHV values achieved by NSGA-II, Knapsack and PageRank on all 9 problem instances. The \textit{mean}, standard deviation ($\sigma$), and best (\textit{max}) values were computed over the 30 independent executions performed for each algorithm. In addition, the table also reports the $p$-value from the Shapiro-Wilk test (\textit{S-W}), the rank from the Friedman test (\textit{Rank}), and the $p$-value from the Friedman test.

\begin{table}[!h]
\centering
\caption{Relative hypervolume values achieved by NSGA-II, Knapsack, and PageRank.}
\small
\setlength{\tabcolsep}{4pt}
\begin{tabular}{ccrrrrrrrrrrr}
\toprule
&  & \multicolumn{3}{c}{normal}& & \multicolumn{3}{c}{high}& & \multicolumn{3}{c}{low}\\
\cline{3-5}\cline{7-9}\cline{11-13}
&  & \multicolumn{1}{r}{data} & \multicolumn{1}{r}{voice} & \multicolumn{1}{r}{video}& & \multicolumn{1}{r}{data} & \multicolumn{1}{r}{voice} & \multicolumn{1}{r}{video}& & \multicolumn{1}{r}{data} & \multicolumn{1}{r}{voice} & \multicolumn{1}{r}{video} \\
\midrule
\multirow{3}{*}{\textit{max}} & NSGA-II & 0.99 & 0.99 & 0.99 & &0.99 & 0.99 & 0.99 && 0.99 & 0.99 & 0.99 \\
					 & Knapsack & 0.82 & 0.85 & 0.95 & &0.8 & 0.84 & 0.96 && 0.83 & 0.86 & 0.95 \\
					 & PageRank & 0.73 & 0.82 & 0.73 & &0.7 & 0.81 & 0.71 & &0.76 & 0.83 & 0.74 \\ 
\midrule
\multirow{3}{*}{\textit{mean}} & NSGA-II & 0.98 & 0.98 & 0.98 & &0.99 & 0.95 & 0.98 & &0.99 & 0.99 & 0.98 \\ 
					 & Knapsack & 0.8 & 0.84 & 0.94 & &0.79 & 0.82 & 0.94 & &0.82 & 0.85 & 0.94 \\ 
					 & PageRank & 0.73 & 0.82 & 0.73 & &0.7 & 0.81 & 0.71 & &0.76 & 0.83 & 0.74 \\ 
\midrule
\multirow{3}{*}{$\sigma$ ($\times10^{-3}$)} & NSGA-II & 3.72 & 2.83 & 3.00 && 2.73 & 176 & 3.79 && 3.74 & 3.65 & 3.33 \\ 
											 & Knapsack & 5.57 & 4.57 & 7.77 && 5.88 & 6.81 & 6.43 && 7.26 & 6.53 & 8.05 \\ 
											 & PageRank & 0 & 0 & 0& & 0 & 0 & 0 && 0 & 0 & 0 \\ 
\midrule
\multirow{3}{*}{\textit{S-W}} & NSGA-II & 0.93 & 0.17 & 0.6& & 0.1 & 0 & 0.25 && 0.3 & 0.11 & 0.35 \\ 
					 & Knapsack & 0.49 & 0.91 & 0.6& & 0.29 & 0.88 & 0.94 && 0.33 & 0.37 & 0.15 \\ 
					 & PageRank & 1 & 1 & 1 & & 1 & 1 & 1 && 1 & 1 & 1 \\ 
\midrule
\multirow{3}{*}{\textit{Rank}} & NSGA-II & 90 & 90 & 90 & & 90 & 88 & 90 && 90 & 90 & 90 \\ 
					 & Knapsack & 60 & 60 & 60 && 60 & 60 & 60 && 60 & 60 & 60 \\ 
					 & PageRank & 30 & 30 & 30 && 30 & 32 & 30 && 30 & 30 & 30 \\ 
\midrule
\multicolumn{2}{c}{$p$-value Friedman} & \scriptsize$<$10$^{-6}$ & \scriptsize$<$10$^{-6}$ & \scriptsize$<$10$^{-6}$ & & \scriptsize$<$10$^{-6}$ & \scriptsize$<$10$^{-6}$ & \scriptsize$<$10$^{-6}$ & & \scriptsize$<$10$^{-6}$ & \scriptsize$<$10$^{-6}$ & \scriptsize$<$10$^{-6}$ \\ 
\bottomrule
\normalsize
\end{tabular}
\label{Tab:RHV}
\end{table}

\begin{table}[!h]
\centering
\caption{NSGA-II improvements over Knapsack and PageRank algorithms.}
\begin{tabular}{lcrrrrrr}
\toprule
 & \multicolumn{1}{l}{} & \multicolumn{ 4}{c}{improvement in QoS (\%)} & \multicolumn{ 2}{c}{improvement in cost (\%)} \\ 
 & \multicolumn{1}{l}{} & \multicolumn{ 2}{c}{cost=\$10000} & \multicolumn{ 2}{c}{cost=\$15000} & \multicolumn{ 2}{c}{QoS=2500} \\
 & \multicolumn{1}{l}{} & \multicolumn{1}{c}{Knapsack} & \multicolumn{1}{c}{Pagerank} & \multicolumn{1}{c}{Knapsack} & \multicolumn{1}{c}{Pagerank} & \multicolumn{1}{c}{Knapsack} & \multicolumn{1}{c}{Pagerank} \\ \midrule
\multicolumn{ 1}{c}{\multirow{3}{*}{normal}} & data & 22.37 & 44.89 & - & - & 31.37 & 39.48 \\ 
 & voice & 19.06 & 29.76 & - & - & 25.34 & 28.93 \\ 
 & video & 2.58 & 49.97 & 5.97 & 24.63 & 6.46 & 31.18 \\ \midrule
\multicolumn{ 1}{c}{\multirow{3}{*}{high}} & data & 24.68 & 52.71 & - & - & 30.17 & 42.4 \\ 
 & voice & 22.11 & 33.82 & - & - & 22.35 & 28.26 \\ 
 & video & 0.83 & 52.55 & 6.42 & 34.13 & 1.84 & 34.18 \\ \midrule
\multicolumn{ 1}{c}{\multirow{3}{*}{low}} & data & 18.44 & 35.61 & - & - & 34.09 & 33.52 \\ 
 & voice & 17.83 & 25.78 & - & - & 33.75 & 25.51 \\ 
 & video & 6.91 & 46.89 & 5.04 & 14.03 & 15.06 & 28.53 \\ \bottomrule
\end{tabular}
\label{Tab:Mejoras}
\end{table}

The results in Table \ref{Tab:RHV} demonstrate that NSGA-II is able to achieve significantly better RHV values than the two baseline heuristics adapted from the literature, both in average and in the best case.
The Friedman rank test allows to state with statistical confidence that NSGA-II is able to outperform both Knapsack and PageRank algorithms in all problem instances ($p$-value $<10^{-6}$ in all comparisons).
This fact suggests that NSGA-II accurately computes fronts that converge towards an ideal Pareto front of the problem while simultaneously maintaining diversity among the set of non-dominated solutions.

Table~\ref{Tab:Mejoras} reports the improvements achieved by NSGA-II over Knapsack and PageRank heuristics. The reported improvements are those achieved when comparing specific (realistic) solutions from the global Pareto fronts computed by each algorithm (i.e., combining the results of all 30 independent executions performed), as usual in the MOEA literature\cite{Coello2002,Deb2001}.

In Table~\ref{Tab:Mejoras}, the improvements regarding each problem objective are evaluated by comparing the values of QoS/cost computed by each studied algorithm when considering a fixed cost/QoS used as a reference value, respectively. The improvements in QoS are measured comparing the values of QoS achieved by each algorithm with a fixed cost of \$10000 (all cost values are expressed in US dollars). This cost is a reasonable value for the budget to invest in order to deploy a RSU infrastructure for VANET in a city scaled area, just like in the case of M\'alaga. In addition, in the scenarios corresponding to video applications, the QoS improvements with a fixed cost of \$15000 are also reported, since the deployment costs for infrastructures that support this type of applications tend to be more expensive. Following a similar approach, the improvements in cost are reported comparing the cost values of solutions achieved by each algorithm with a fixed QoS of 2500.

In any case, since the proposed algorithm solves the problem following a multiobjective approach, a human decision-maker will always have the last word about the solution to implement: the decision-maker from the City Council will be in charge of selecting one solution of the set of non-dominated ones for the real implementation over the city. 

The results in Table \ref{Tab:Mejoras} clearly state that NSGA-II is able to improve the solutions computed by both Knapsack and PageRank algorithms in all scenarios. In terms of QoS, the maximum improvements are of 24.68\% over Knapsack and 52.71\% over PageRank, both in the instance with high traffic and a data application.
The improvements of costs are of up to 34.09\% over Knapsack (instance with low traffic and a data application) and 39.48\% over PageRank (instance with normal traffic and a data application).

\begin{figure}[!h]
\centering
\hspace{-1.1\baselineskip}

\begin{subfigure}{.33\textwidth}
\centering
\includegraphics[width=\textwidth]{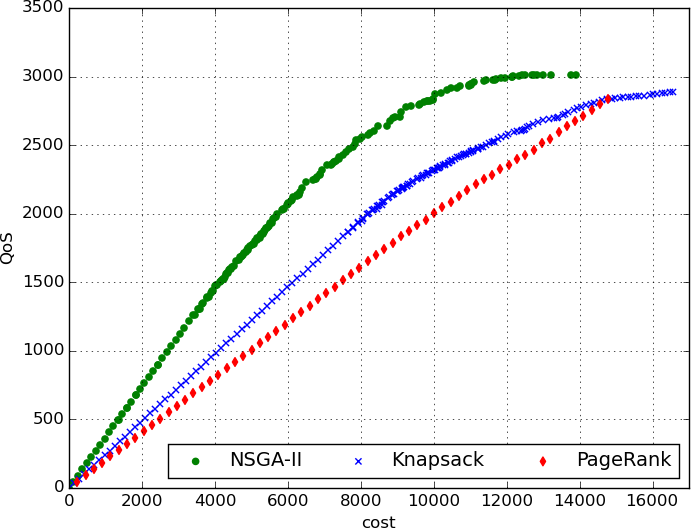}
\caption{Normal traffic, data app.}
\label{Fig:PF_normal_data}
\end{subfigure}
\begin{subfigure}{.33\textwidth}
\centering
\includegraphics[width=\linewidth]{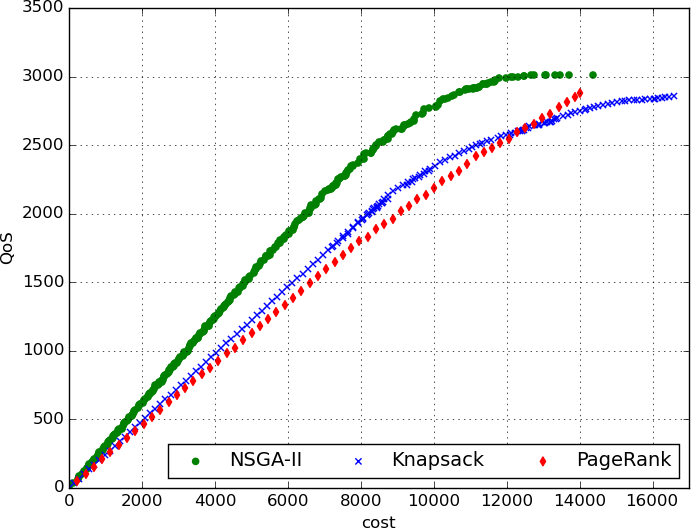}
\caption{Normal traffic, voice app.}
\label{Fig:PF_normal_voice}
\end{subfigure}
\begin{subfigure}{.33\textwidth}
\centering
\includegraphics[width=\linewidth]{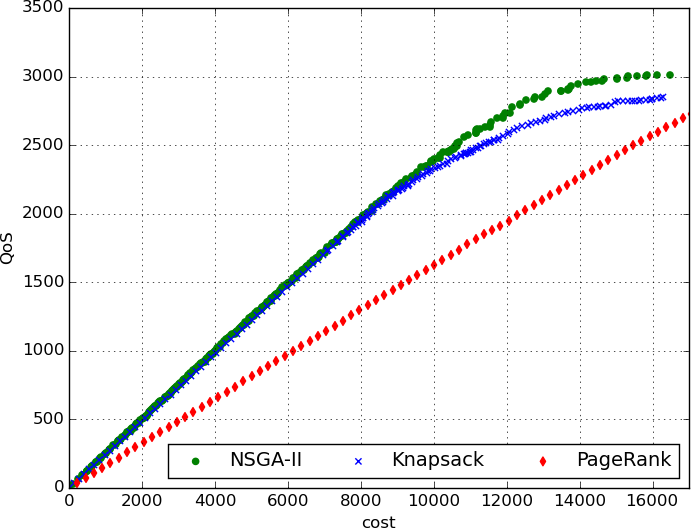}
\caption{Normal traffic, video app.}
\label{Fig:PF_normal_video}
\end{subfigure}
\hspace{-1.1\baselineskip}
\vspace{0.4cm}

\begin{subfigure}{.33\textwidth}
\centering
\includegraphics[width=\linewidth]{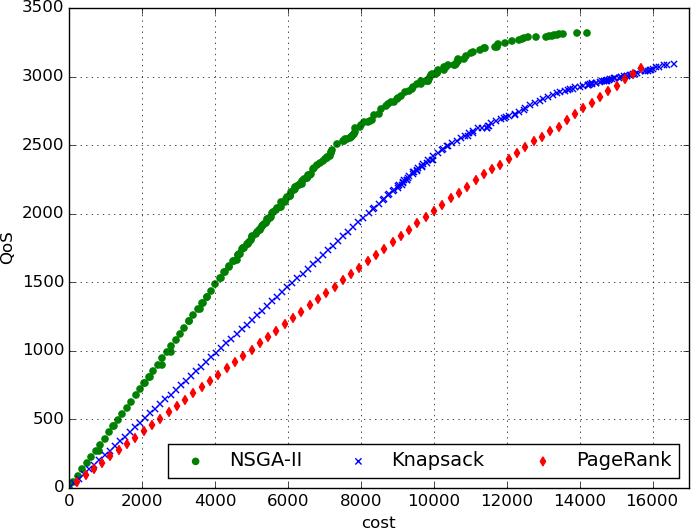}
\caption{High traffic, data app.}
\label{Fig:PF_high_data}
\end{subfigure}
\begin{subfigure}{.33\textwidth}
\centering
\includegraphics[width=\linewidth]{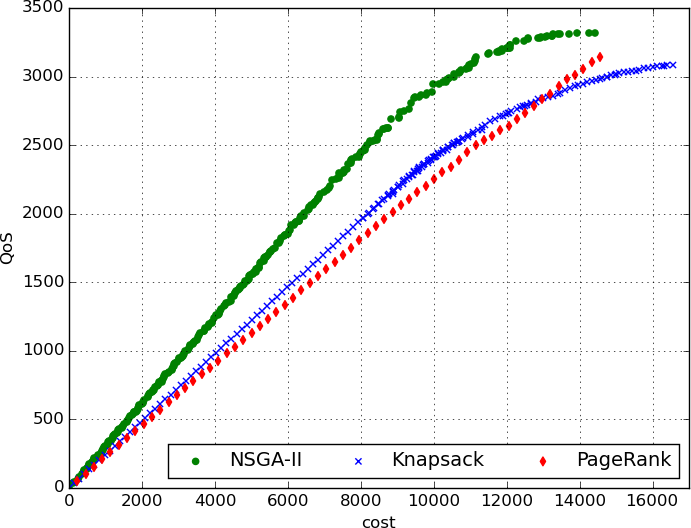}
\caption{High traffic, voice app.}
\label{Fig:PF_high_voice}
\end{subfigure}
\begin{subfigure}{.33\textwidth}
\centering
\includegraphics[width=\linewidth]{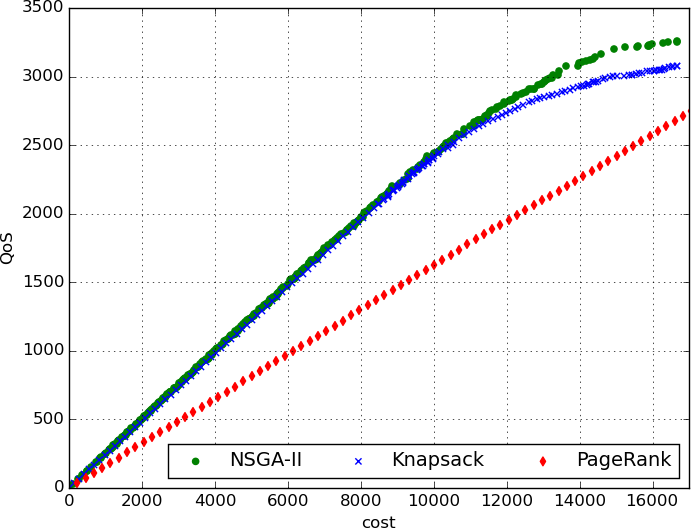}
\caption{High traffic, video app.}
\label{Fig:PF_high_video}
\end{subfigure}
\hspace{-1.1\baselineskip}
\vspace{0.4cm}

\begin{subfigure}{.33\textwidth}
\centering
\includegraphics[width=\linewidth]{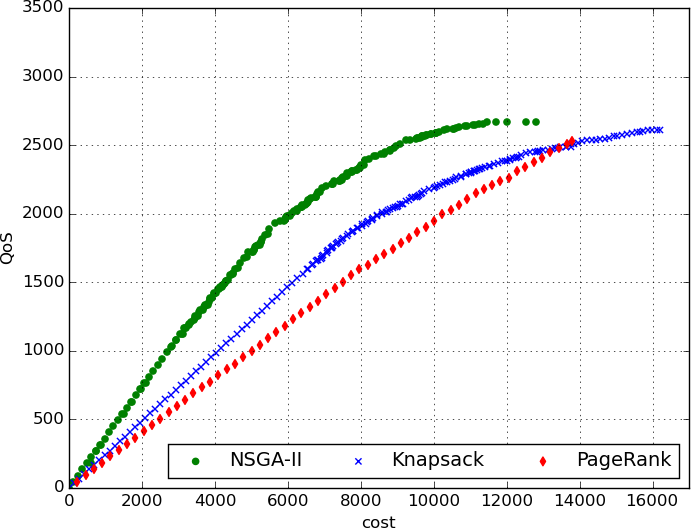}
\caption{Low traffic, data app.}
\label{Fig:PF_low_data}
\end{subfigure}
\begin{subfigure}{.33\textwidth}
\centering
\includegraphics[width=\linewidth]{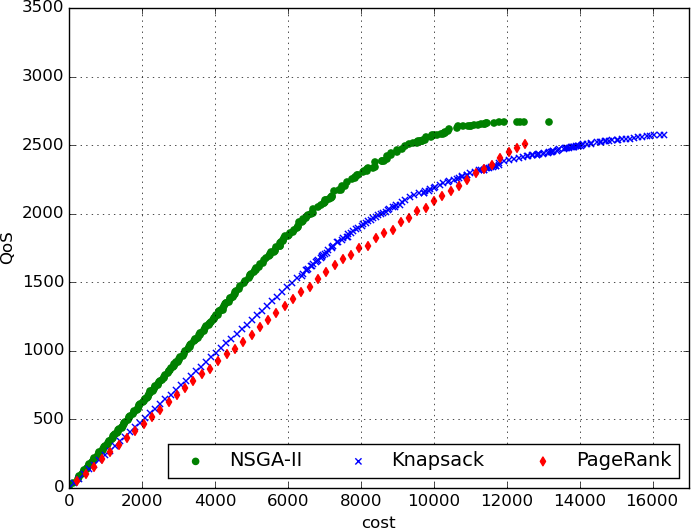}
\caption{Low traffic, voice app.}
\label{Fig:PF_low_voice}
\end{subfigure}
\begin{subfigure}{.33\textwidth}
\centering
\includegraphics[width=\linewidth]{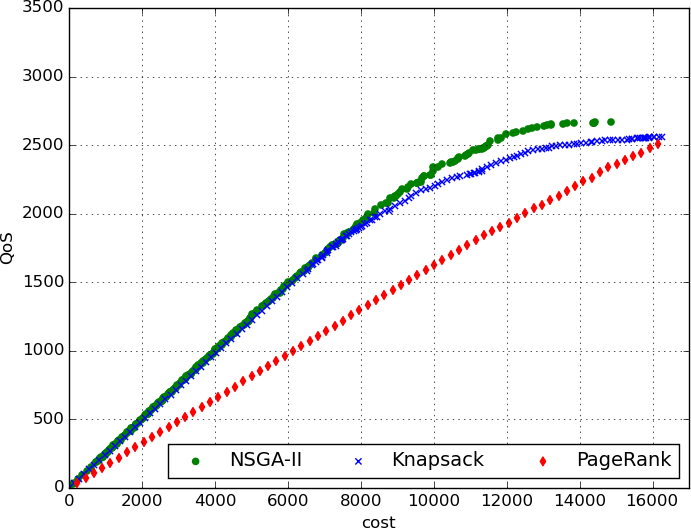}
\caption{Low traffic, video app.}
\label{Fig:PF_low_video}
\end{subfigure}
\hspace{-1.1\baselineskip}
\vspace{0.4cm}

\caption{Global Pareto fronts.}
\label{Fig:global_pf}
\end{figure}

In order to provide a better insight into the main advantages of the proposed NSGA-II algorithm, Figure~\ref{Fig:global_pf} shows the global Pareto fronts achieved by each algorithm on each problem instance.
It can be observed that NSGA-II is able to compute accurate Pareto fronts with good convergence and diversity properties. The improvements of NSGA-II on both QoS and cost are clear for data and voice applications, in all three traffic patterns studied. When considering a more demanding application (video), the improvements of NSGA-II over Knapsack are observed for large values of QoS (i.e., QoS $> 2500$). The PageRank algorithm was consistently the worst method to solve the RSU-DP in all problem instance studied.

\vspace{-6pt}
\section{Conclusions and future work}
\label{Sec:Conc}
\vspace{-2pt}

This article reports the application of a multiobjective evolutionary approach to solve the problem of locating roadside infrastructure for vehicular networks over realistic urban areas.

A multiobjective formulation of the problem was introduced, considering the QoS and cost objectives. A specific NSGA-II evolutionary algorithm was designed, by including a problem-related encoding and ad-hoc mutation operators to explore the set of possible locations. A parallel model was applied in order to efficiently perform the evaluation of solutions in the proposed MOEA.

The NSGA-II algorithm was evaluated on a city-scaled area using the real map and up-to-date (year 2015) traffic information from the city of M\'alaga, in Spain. The problem instances were built considering three different types of commercial antennas and a set of realistic VANET parameterizations that could ease the use of demanding applications related to data, voice, and video communications. The scenarios used for the experimental evaluation considered three different traffic patterns and three different types of applications, each one with different infrastructure requirements.

Two heuristics were implemented to solve the problem, based on related works from the literature: a Randomized Knapsack algorithm and a constructive PageRank heuristic. These traditional methods for RSU planning were used as a reference baseline to compare the solutions computed by the proposed NSGA-II algorithm. In the experiments performed, the proposed MOEA has shown good problem solving capabilities, computing accurate Pareto fronts for the problem. NSGA-II also allows improving over the two baseline heuristics, regarding the multiobjective optimization metrics evaluated and the two problem objectives.
 
According to the results from the experimental analysis, NSGA-II was able to consistently compute better results than the baseline heuristics. In the best case, NSGA-II was able to outperform the Randomized Knapsack algorithm up to 24.68\% and the PageRank heuristic up to 52.71\% in terms of QoS. The improvements in cost were up to 34.09\% and 39.48\% over Knapsack and PageRank, correspondingly. The computed Pareto fronts indicate that NSGA-II provides better and more robust solutions to the problem, especially for those scenarios considering data and voice applications. The experimental analysis also demonstrates that when considering a more realistic scenario (including real applications, real traffic information, etc.) the evolutionary approach is able to achieve larger improvements over the heuristics than those reported in previous related works.

The main lines for future work are related to extend the experimental analysis to consider other real areas from cities in other countries. Right now, we are working on building a real RSU-DP scenario for VANETs in the city of Montevideo, Uruguay, using real data from the local authorities and GPS information from the public transport. We also plan to extend the problem formulation to consider additional information from important events in vehicular networks (such as accidents, traffic jams, etc.) in order to model a more realistic scenario for the problem.

\vspace{-6pt}
\section*{Acknowledgments}
\vspace{-2pt}

The work of R. Massobrio and S. Nesmachnow has been partially funded by ANII and PEDECIBA, Uruguay.
The work of J. Toutouh and E. Alba has been partially funded by the Spanish MINECO and FEDER project TIN2014-57341-R.

\vspace{-6pt}
\section*{Contributions}
\vspace{-2pt}
Renzo Massobrio and Jamal Toutouh carried out most of the experimental study, including programming the multiobjective evolutionary approach and the heuristics to solve the problem, and performing the VANET experiments to build the real scenario studied in the experimental analysis. They also participated in the writing of the paper. Sergio Nesmachnow worked in the design and implementation of the software methods, the design and evaluation of the case study in M\'alaga, and is responsible of the writing of the paper. Enrique Alba has contributed in the revision of the manuscript and writing of the paper. The international cooperation was coordinated by S. Nesmachnow (U.~de la Rep\'ublica) and J. Toutouh (U.~of M\'alaga). S.~Nesmachnow and E.~Alba were responsible for the funds dedicated to the international cooperation.

\bibliography{paper}

\begin{thebibliography}{10}

\bibitem{Alba2013}
E.~Alba, G.~Luque, and S.~Nesmachnow.
\newblock Parallel metaheuristics: Recent advances and new trends.
\newblock {\em International Transactions in Operational Research},
  20(1):1--48, 2013.

\bibitem{Aslam2012}
B.~Aslam, F.~Amjad, and C.~Zou.
\newblock Optimal roadside units placement in urban areas for vehicular
  networks.
\newblock In {\em IEEE Symposium on Computers and Communications}, pages
  423--429, July 2012.

\bibitem{BFM97}
T.~B\"{a}ck, D.~Fogel, and Z.~Michalewicz, editors.
\newblock {\em Handbook of evolutionary computation}.
\newblock Oxford University Press, 1997.

\bibitem{Brahim2014}
Mohamed~Ben Brahim, Wassim Drira, and Fethi Filali.
\newblock Roadside units placement within city-scaled area in vehicular ad-hoc
  networks.
\newblock In {\em 3$^{rd}$ International Conference on Connected Vehicles and
  Expo}, pages 1--7. IEEE, 2014.

\bibitem{Campolo2015}
C.~Campolo, A.~Molinaro, and R.~Scopigno, editors.
\newblock {\em Vehicular ad hoc Networks - Standards, Solutions, and Research}.
\newblock Springer, 2015.

\bibitem{Cavalcante2012}
E.~Cavalcante, A.~Aquino, G.~Pappa, and A.~Loureiro.
\newblock Roadside unit deployment for information dissemination in a {VANET}:
  An evolutionary approach.
\newblock In {\em 14$^{th}$ Genetic and Evolutionary Computation Conference},
  pages 27--34, 2012.

\bibitem{cetacea}
{Cetacea Wireless Solutions Company shop}.
\newblock Online \texttt\footnotesize{https://shop.cetacea.com/}.
\newblock Retrieved December 2015.

\bibitem{Chantaksinopas2012}
I.~Chantaksinopas, W.~Lee, A.~Prayote, and P.~Oothongsap.
\newblock Delay-sensitive applications in {VANET} and seamless connectivity:
  The limitation of {UMTS} network.
\newblock {\em International Journal Computer Science \& Network Security},
  12:54--61, 2012.

\bibitem{Cheng2013}
H.~Cheng, X.~Fei, A.~Boukerche, A.~Mammeri, and M.~Almulla.
\newblock A geometry-based coverage strategy over urban {VANETs}.
\newblock In {\em 10$^{th}$ ACM Symposium on Performance Evaluation of Wireless
  Ad Hoc, Sensor, \& Ubiquitous Networks}, pages 121--128, 2013.

\bibitem{Coello2002}
C.~Coello, D.~Van~Veldhuizen, and G.~Lamont.
\newblock {\em Evolutionary algorithms for solving multi-objective problems}.
\newblock Kluwer, New York, 2002.

\bibitem{Deakin2013}
M.~Deakin.
\newblock {\em Smart Cities: Governing, Modelling, and Analysing the
  Transition}.
\newblock Routledge, 2013.

\bibitem{Deb2001}
K.~Deb.
\newblock {\em Multi-Objective Optimization using Evolutionary Algorithms}.
\newblock J. Wiley \& Sons, Chichester, 2001.

\bibitem{deb2002}
K.~Deb, A.~Pratap, S.~Agarwal, and T.~Meyarivan.
\newblock A fast and elitist multiobjective genetic algorithm: {NSGA-II}.
\newblock {\em IEEE Transactions on Evolutionary Computation}, 6(2):182--197,
  2002.

\bibitem{Falcocchio2015}
J.~Falcocchio and H.~Levinson.
\newblock {\em Road Traffic Congestion: A Concise Guide}.
\newblock Springer Tracts on Transportation and Traffic. Springer
  International, 2015.

\bibitem{Glover1986}
F.~Glover.
\newblock Future paths for integer programming and links to artificial
  intelligence.
\newblock {\em Computers and Operations Research}, 13(5):533--549, 1986.

\bibitem{Goldberg89}
D.~Goldberg.
\newblock {\em Genetic algorithms in search, optimization, and machine
  learning}.
\newblock Addison Wesley, New York, 1989.

\bibitem{Hartenstein2010}
H.~Hartenstein, K.~Laberteaux, and I.~Ebrary.
\newblock {\em {VANET: vehicular applications and inter-networking
  technologies}}.
\newblock Wiley Online Library, 2010.

\bibitem{Jiang2008}
D.~Jiang and L.~Delgrossi.
\newblock {IEEE} 802.11p: Towards an international standard for wireless access
  in vehicular environments.
\newblock In {\em IEEE Vehicular Technology Conference}, pages 2036--2040,
  2008.

\bibitem{Langville2011}
A.~Langville and C.~Meyer.
\newblock {\em Google's PageRank and beyond: The science of search engine
  rankings}.
\newblock Princeton University Press, 2011.

\bibitem{Liang2012}
Y.~Liang, H.~Liu, and D.~Rajan.
\newblock Optimal placement and configuration of roadside units in vehicular
  networks.
\newblock In {\em IEEE 75$^{th}$ Vehicular Technology Conference}, pages 1--6.
  IEEE, 2012.

\bibitem{Lochert2008}
C.~Lochert, B.~Scheuermann, C.~Wewetzer, A.~Luebke, and M.~Mauve.
\newblock Data aggregation and roadside unit placement for a {VANET} traffic
  information system.
\newblock In {\em 5$^{th}$ ACM International Workshop on Vehicular
  Inter-Networking}, pages 58--65, 2008.

\bibitem{Martello1990}
S.~Martello and P.~Toth.
\newblock {\em Knapsack Problems: Algorithms and Computer Implementations}.
\newblock John Wiley \& Sons, Inc., New York, NY, USA, 1990.

\bibitem{Massobrio2015b}
R.~Massobrio, S.~Bertinat, S.~Nesmachnow, J.~Toutouh, and E.~Alba.
\newblock Smart placement of {RSU} for vehicular networks using multiobjective
  evolutionary algorithms.
\newblock In {\em Latin America Conference on Computational Intelligence},
  pages 1--6, 2015.

\bibitem{Massobrio2015}
R.~Massobrio, J.~Toutouh, and S.~Nesmachnow.
\newblock A multiobjective evolutionary algorithm for infrastructure location
  in vehicular networks.
\newblock In {\em 7$^{th}$ European Symposium on Computational Intelligence and
  Mathematics}, pages 1--6, 2015.

\bibitem{Mendes2009}
S.~Mendes, J.~G{\'{o}}mez-Pulido, M.~Vega-Rodr{\'{\i}}guez,
  J.~S{\'{a}}nchez-P{\'{e}}rez, Y.~S{\'{a}}ez, and P.~Isasi.
\newblock The radio network design optimization problem.
\newblock In {\em Biologically-Inspired Optimisation Methods}, pages 219--260.
  Springer Science + Business Media, 2009.

\bibitem{webmovilidad}
{\'Area de Movilidad, Ayuntamiento de M\'alaga}.
\newblock Online http://movilidad.malaga.eu/.
\newblock Retrieved December 2015.

\bibitem{Nesmachnow2010}
S.~Nesmachnow.
\newblock Computaci\'on cient\'ifica de alto desempe\~no en la {Facultad de
  Ingenier\'ia, Universidad de la Rep\'ublica}.
\newblock {\em Revista de la Asociaci\'on de Ingenieros del Uruguay},
  61:12--15, 2010.

\bibitem{Nesmachnow2014}
S.~Nesmachnow.
\newblock An overview of metaheuristics: accurate and efficient methods for
  optimisation.
\newblock {\em International Journal of Metaheuristics}, 3(4):320--347, 2014.

\bibitem{ns-2}
{The Network Simulator ns-2}.
\newblock Online http://www.isi.edu/nsnam/ns.
\newblock Retrieved December 2015.

\bibitem{Patil2013}
P.~Patil and A.~Gokhale.
\newblock Voronoi-based placement of road-side units to improve dynamic
  resource management in {VANETs}.
\newblock In {\em International Conference on Collaboration Technologies and
  Systems}, pages 389--396, 2013.

\bibitem{Reis2014}
A.~Reis, S.~Sargento, F.~Neves, and O.~Tonguz.
\newblock Deploying roadside units in sparse vehicular networks: What really
  works and what does not.
\newblock {\em IEEE Transactions on Vehicular Technology}, 63(6):2794--2806,
  2014.

\bibitem{Saunders1999}
S.~Saunders and A.~Aragon.
\newblock {\em Antennas and Propagation for Wireless Communication Systems}.
\newblock Wiley, New York, NY, USA, 1999.

\bibitem{Trullols2010}
O.~Trullols, M.~Fiore, C.~Casetti, C.~Chiasserini, and J.~Ordinas.
\newblock Planning roadside infrastructure for information dissemination in
  intelligent transportation systems.
\newblock {\em Computer Communications}, 33(4):432--442, 2010.

\bibitem{Chengyuan2014}
C.~Wang, X.~Li, F.~Li, and H.~Lu.
\newblock A mobility clustering-based roadside units deployment for {VANET}.
\newblock In {\em 16$^{th}$ Asia-Pacific Network Operations and Management
  Symposium}, pages 1--6, 2014.

\bibitem{White2011}
D.~White.
\newblock Software review: the {ECJ} toolkit.
\newblock {\em Genetic Programming and Evolvable Machines}, 13(1):65--67, 2011.

\end{thebibliography}
\bibliographystyle{plain}

\end{document}